\documentclass[twoside,11pt]{article}
\usepackage{blindtext}
\usepackage{url}
\usepackage{times}
\usepackage{epsfig}
\usepackage{graphicx}
\usepackage{amsmath}
\usepackage{amssymb}
\usepackage{algorithmic}
\usepackage[linesnumbered,ruled,vlined]{algorithm2e}
\usepackage{multirow}
\usepackage{booktabs}
\usepackage{siunitx}
\usepackage{bm}
\usepackage{amsfonts}
\usepackage{caption} 
\usepackage{color}
\usepackage{graphicx}

\usepackage{jmlr2e}

\usepackage{lastpage}
\jmlrheading{23}{2022}{1-\pageref{LastPage}}{1/21; Revised 5/22}{9/22}{21-0000}{Author One and Author Two}

\ShortHeadings{Sample JMLR Paper}{One and Two}
\firstpageno{1}

\begin{document}

\title{Decision Boundary Optimization-Informed Domain Adaptation}

\author{\name Lingkun Luo \email lolinkun@gmail.com \\
       \addr Department of Information and Control\\
       Shanghai Jiao Tong University\\
       800 Dongchuan Road, Shanghai, China
       \AND
       \name Shiqiang Hu \email sqhu@sjtu.edu.cn \\
       \addr Department of Information and Control\\
       Shanghai Jiao Tong University\\
       800 Dongchuan Road, Shanghai, China
       \AND 
       \name Jie Yang \email jieyang@sjtu.edu.cn \\
       \addr Department of Electronic, Information and Electrical Engineering\\
       Shanghai Jiao Tong University\\
       800 Dongchuan Road, Shanghai, China
       \AND
       \name ~Liming~Chen \email liming.chen@ec-lyon.fr \\
       \addr Department of Mathematics and Informatics\\
       École Centrale de Lyon\\
        69134 Écully, France}

\editor{My editor}

\maketitle

\begin{abstract}
Maximum Mean Discrepancy (\textbf{MMD})  is  widely used in a number of domain adaptation (\textbf{DA}) methods and shows its effectiveness in aligning data distributions across domains.  However, in previous \textbf{DA} research, MMD-based \textbf{DA} methods  focus mostly on distribution alignment, and ignore to optimize the decision boundary for classification-aware DA, thereby falling short in  reducing the \textbf{DA} upper error bound. In this paper, we propose a strengthened \textbf{MMD} measurement, namely, \textit{D}ecision \textit{B}oundary optimization-informed MMD (\textbf{DB-MMD}), which enables \textbf{MMD}  to carefully take into account the decision boundaries, thereby simultaneously optimizing the distribution alignment and cross-domain classifier  within a hybrid framework, and leading to a \textit{theoretical bound} guided \textbf{DA}. We further seamlessly embed the proposed \textbf{DB-MMD} measurement into several popular \textbf{DA} methods, \textit{e.g.}, \textbf{MEDA}, \textbf{DGA-DA},   to demonstrate its effectiveness  \textit{w.r.t} different experimental settings. We carry out comprehensive experiments using 8 standard \textbf{DA} datasets. The experimental results show that the \textbf{DB-MMD} enforced \textbf{DA} methods  improve their baseline models using plain vanilla \textbf{MMD}, with a   margin that  can be as high as $9.5$.
\end{abstract}


\section{Introduction}

Supervised learning, \textit{e.g.},  deep learning, has witnessed great progress in both theory and practice in recent years. Its basic assumption assumes that the training and testing data are drawn from a same distribution. However, in real-life applications it frequently happens that a predictor learned from well labeled data in a source domain can not be applied to testing data in a target domain, because of the well-known  \textit{domain shift} phenomenon. Due to factors as diverse as sensor difference, viewpoint variations, and lighting changes, \textit{etc},  \cite{DBLP:journals/csur/LuLHWC20,pan2010survey,7078994,DBLP:journals/tnn/ShaoZL15}, testing data in the target domain can be very different from the learning data in the source domain, thereby requiring huge laborious human efforts to label data of the target domain for the purpose of retraining the learned model.

Unsupervised domain adaptation (\textbf{UDA}) aims at addressing the former issue and developing methods and techniques, which significantly improve the performance of traditional machine learning techniques within the cross-domain scenario and  make use of a model trained on the well labeled source domain$({{\cal D}_{\cal S}})$ for direct application to the unlabeled target domain$({{\cal D}_{\cal T}})$ regardless of the existing \textit{domain shift}.  Specifically, the rationale of \textbf{UDA} in solving cross-domain tasks can be well explained via  the cornerstone theoretical result in \textbf{DA}\cite{ben2010theory,kifer2004detecting}, which states the error bound of a learned hypothesis $h$ on the target domain:  


\begin{equation}\label{eq:bound}
			\begin{array}{l}
			{e_{\cal T}}(h) \le {e_{\cal S}}(h) + {d_{\cal H}}({{\cal D}_{\cal S}},{{\cal D}_{\cal T}})+ \min \left\{ {{{\cal E}_{{{\cal D}_{\cal S}}}}\left[ {\left| {{f_{\cal S}}({\bf{x}}) - {f_{\cal T}}({\bf{x}})} \right|} \right],{{\cal E}_{{{\cal D}_{\cal T}}}}\left[ {\left| {{f_{\cal S}}({\bf{x}}) - {f_{\cal T}}({\bf{x}})} \right|} \right]} \right\}
			\end{array}
	\end{equation}

where the performance ${e_{\cal T}}(h)$ of a hypothesis $h$ on the target domain in the left-hand is bounded by the following three terms in the right-hand:

	\begin{itemize}
	\item \textbf{Term.1}: ${e_{\cal S}}(h)$ denotes the classification error on the source domain;
	
	\item \textbf{Term.2}: ${{d_{\cal H}}({{\cal D}_{\cal S}},{{\cal D}_{\cal T}})}$ measures the $\mathcal{H}$\emph{-divergence}\cite{kifer2004detecting} between two distributions ($\mathcal{D_S}$, $\mathcal{D_T}$);

	\item \textbf{Term.3}:  the last term characterizes the difference in labeling functions across the two domains;	
	
    \end{itemize}

In light of this theoretic error bound, it can be observed that popular \textbf{DA} approaches have  either focused  on \textit{distribution alignment} to reduce \textbf{Term.2}; or \textit{classifier optimization} to decrease \textbf{Term.1\&3} for theory grounded functional learning:

	\begin{itemize}
	\item \textbf{Distribution Alignment} based \textbf{DA (DA-DA)}: the main research goal of \textbf{DA-DA} approaches is to align the cross-domain data distributions through different statistical measurements, \textit{e.g.}, Bregman Divergence \cite{4967588},  Wasserstein distance \cite{courty2017joint, courty2017optimal}, Maximum Mean Discrepancy (\textbf{MMD})\cite{DBLP:conf/nips/GrettonBRSS06}, \textit{etc}.  \textbf{MMD} based \textbf{DA} has received much attention so far in the research community  \cite{pan2011domain,wang2018visual,DBLP:conf/icml/LongZ0J17,long2013transfer} thanks to its solid theoretical foundation and simplicity.
	
	\item \textbf{Classifier Optimization} based \textbf{DA (CO-DA)}: these approaches embrace the classifier optimization strategies to simultaneously guarantee the source error minimization \cite{li2021cross,yang2020fda,lee2021dranet} and an alignment of cross-domain classifiers  \cite{DBLP:conf/cvpr/SaitoWUH18,liang2021domain}.
    \end{itemize}

\begin{figure}[h!]
	\centering
	\includegraphics[width=0.7\linewidth]{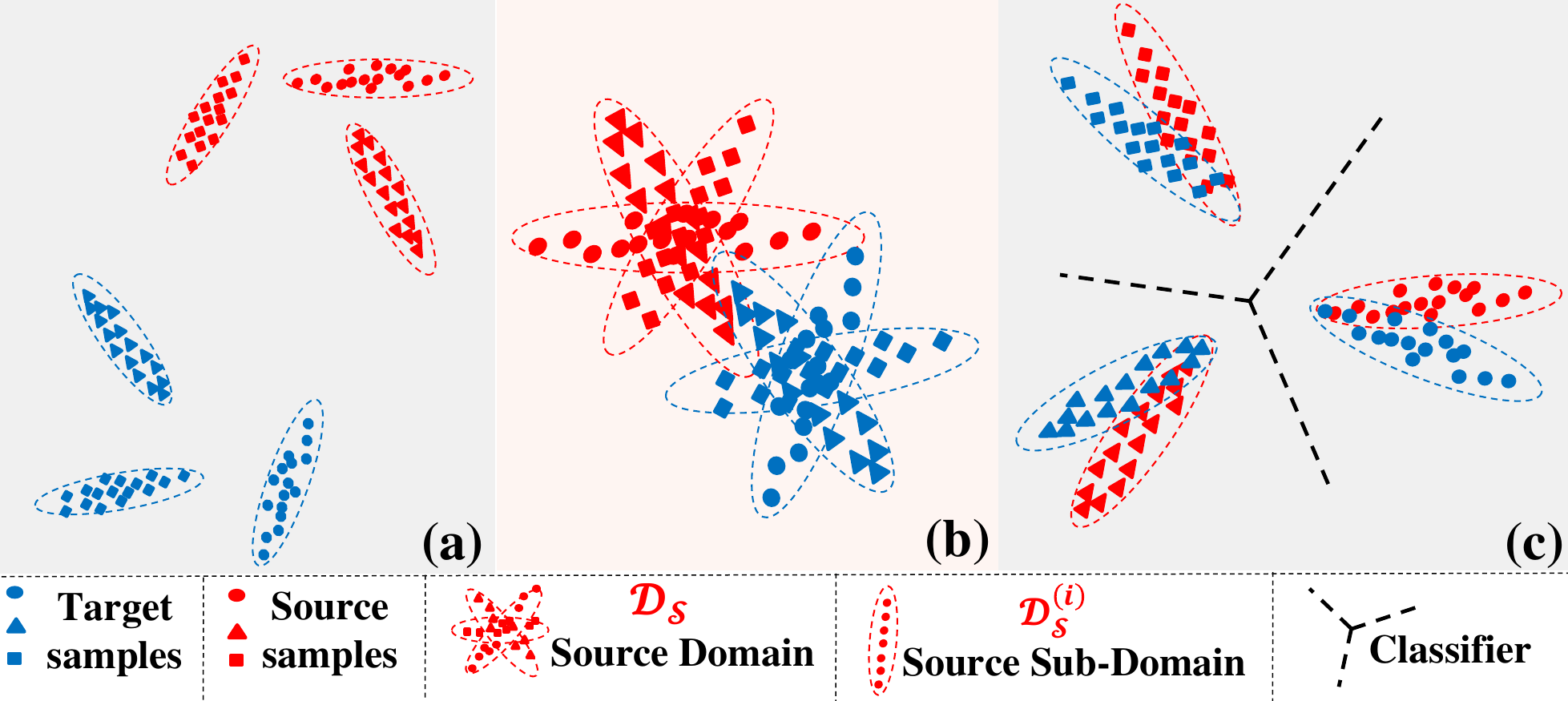}

	\caption {Fig.\ref{fig:1}.(a) shows that the source domain and the target domain samples depict a large domain divergence in the original feature space. Fig.\ref{fig:1}.(b) highlights that  \textbf{distribution alignment}-based \textbf{DA}   drags close the domains and the sub-domains but tends to ignore to optimize the decision boundary for yielding the \textbf{classifier optimization} ensured functional learning as required in Fig.\ref{fig:1}.(c).}
		\label{fig:1}
\end{figure}     
    
Ideally, an effective \textbf{DA} method should take virtue of both \textbf{DA-DA} and \textbf{CO-DA} to comprehensively minimize the error bound, in a such way that not only the \textit{distribution alignment} theoretically guarantees the domain shift reduction, but also  facilitate  the \textit{classifier optimization} to solve the cross-domain tasks, \textit{i.e.}, image classification, semantic segmentation, \textit{etc}. However, in \textbf{DA-DA}, popular statistic measurements (\textit{e.g.}, \textbf{MMD}), tend to be more focused  on distribution divergence reduction than the decision boundary awareness, thereby  unable to proactively serve for the next round \textit{classifier optimization}. As visualized in Fig.\ref{fig:1}.(b), the \textbf{MMD} enforced \textit{distribution alignment} merely brings closer the cross domains/sub-domains, but ignores to optimize the decision boundary for  the \textit{classifier optimization} required functional learning as shown in Fig.\ref{fig:1}.(c).

As a result, some recently proposed \textbf{DA} methods \cite{ganin2016domain,tzeng2017adversarial,pei2018multi,pmlr-v80-hoffman18a,kim2019unsupervised,li2021bi} hybridize the \textit{distribution alignment} and \textit{classifier optimization} within a unified optimization framework to achieve remarkable progress in terms of performance. However, these approaches still do not explicitly enable across domain data distribution alignment with decision boundary awareness,  thereby increasing the burden of the additional classifier regularization. To address this issue, we propose in this paper a novel statistic measurement, namely, \textbf{D}ecision \textbf{B}oundary Optimization-Informed \textbf{MMD} (\textbf{DB-MMD}), which provides a harmonious blending of the \textit{distribution alignment} and \textit{classifier optimization} to draw the best of the two worlds, thereby improving \textbf{MMD} based  \textbf{DA} baselines with theoretical guarantees.

\begin{figure}[h!]
	\centering
	\includegraphics[width=0.8\linewidth]{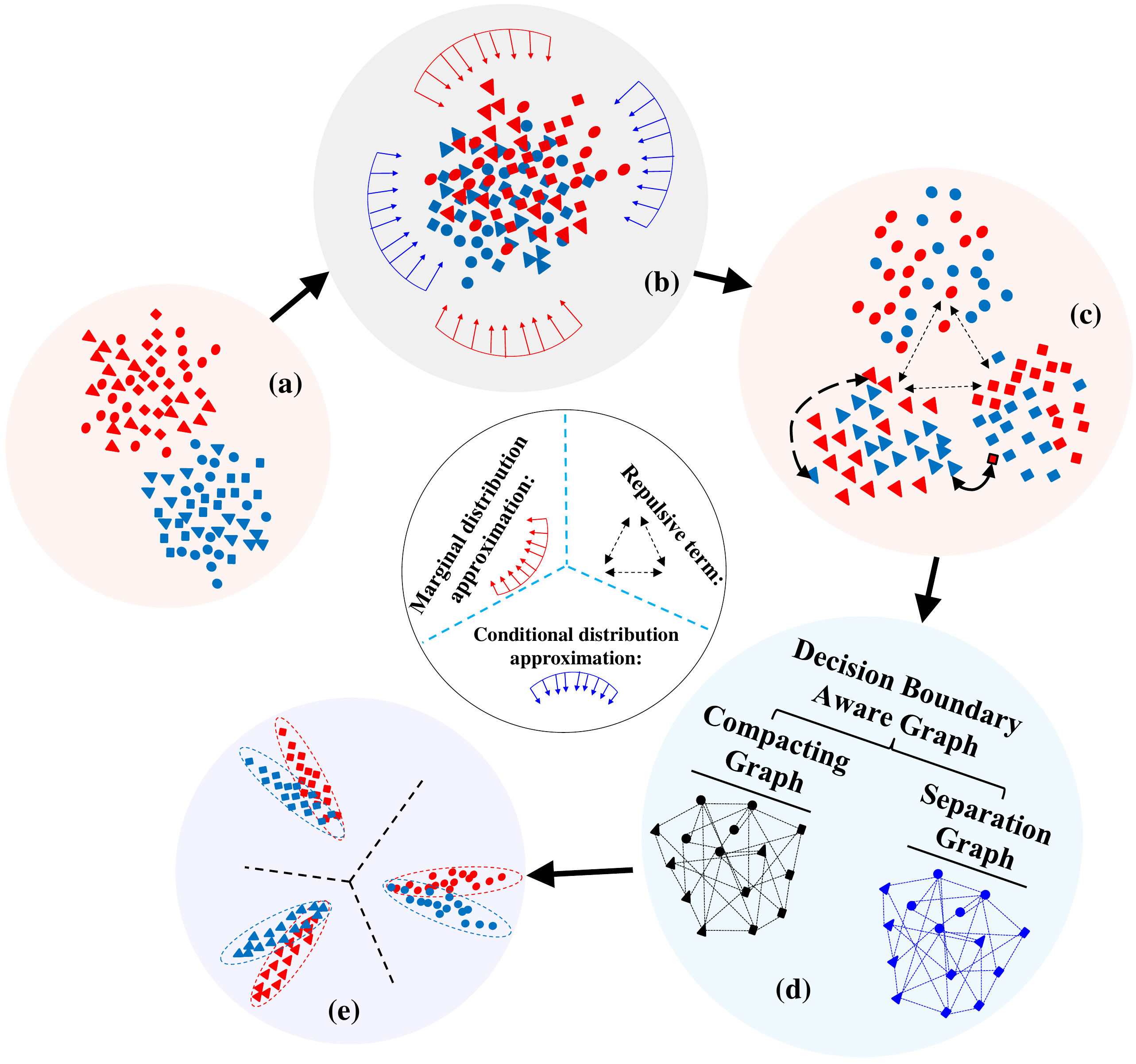}

	\caption {Illustration of the proposed decision boundary optimization-informed \textbf{DA} (\textbf{DB-DA}). Fig.\ref{fig:2} (a): the original source and target domain distributions; 	Fig.\ref{fig:2} (b,c) illustrate \textbf{DB-DA} aligning cross-domain distributions closely yet discriminatively by using MMD. Fig.\ref{fig:2} (d) shows the proposed \textbf{DA} aware of decision boundary  through the specifically designed '\textit{compacting graph}' and '\textit{separation graph}'; Fig.\ref{fig:2} (e) illustrates the achieved latent joint subspace where both marginal and class conditional data distributions are aligned discriminatively and the decision boundaries are clearly optimized.}
		\label{fig:2}
\end{figure}

Specifically, Fig.\ref{fig:2}  depicts the overall framework of the \textbf{DB-MMD} design process. As visualized in Fig.\ref{fig:2}.(a), the existing \textit{domain shift}  hinders the learned model on the source domain ($\mathcal{D_S}$) to be effective on the target domain ($\mathcal{D_T}$).

\begin{itemize}
	\item \textbf{Distribution alignment}: in Fig.\ref{fig:2}.(b), the designed \textbf{DB-MMD} explicitly brings close the cross domains/sub-domains to reduce the \textit{domain shift}, thereby reducing \textbf{Term.2}. In unsupervised \textbf{DA}, the definition of sub-domains in the target domain requires a base classifier,\textit{ e.g.}, Nearest Neighbor (NN),  to attribute  pseudo labels for samples in ${{\cal D}_{\cal T}}$.

 	\end{itemize}

	\begin{itemize}
	\item \textbf{Discriminativeness across sub-domains}: in Fig.\ref{fig:2}.(c), the designed \textbf{DB-MMD} further explores data discriminativeness to explicitly drag away the differently labeled sub-domains, thereby potentially reducing the risk of misclassification on the source domain and decreasing \textbf{Term.1}.
	    \end{itemize}

	\begin{itemize}
	\item \textbf{Decision boundary awareness}: using the specifically designed decision boundary aware graph as depicted in Fig.\ref{fig:2}.(d), the basic \textbf{MMD} measurement is strengthened to become \textbf{D}ecision \textbf{B}oundary optimization-informed \textbf{MMD} (\textbf{DB-MMD}), which compacts intra-class samples while separating inter-class samples for the decision boundary aware \textbf{DA} as shown in Fig.\ref{fig:2}.(e) and reduces \textbf{Term.3}.

    \end{itemize}

Therefore, the proposed novel \textbf{MMD} measurement, namely, \textbf{D}ecision \textbf{B}oundary optimization-informed \textbf{MMD} (\textbf{DB-MMD}),  enables  simultaneous optimization of all the three terms of the \textit{hypothesis error bound} on the target domain (Eq.(\ref{eq:bound})) and makes it possible for effective domain adaptation.

In summary, the main contributions of this paper are as follows:

	\begin{itemize}
		\item A novel statistic measurement, \textit{i.e.}, \textbf{D}ecision \textbf{B}oundary optimization-informed \textbf{MMD} (\textbf{DB-MMD}) is proposed, in which a \textit{decision boundary aware graph} is specifically designed to make the \textbf{MMD} regularized \textit{distribution alignment} aware of decision boundaries so as to proactively serve  \textit{classifier optimization}, thus leading to the \textbf{D}ecision \textbf{B}oundary optimization-informed \textbf{D}omain \textbf{A}daptation (\textbf{DB-DA}).

	\item By embedding the proposed \textbf{DB-MMD} into different MMD-based unsupervised \textbf{DA} baselines, we readily obtain  their strengthened \textbf{DA} counterparts, which consistently improve their baselines across 8 standard \textbf{DA} databases, thereby demonstrating the effectiveness and versatility of the proposed \textbf{DB-MMD}.
    \end{itemize}

\section{Related Work}
\label{Related Work}

Because domain shift frequently occurs in real-life applications,  \textbf{DA} has attracted a lot of attention from the research communities and recorded a number of original and  effective methods \cite{DBLP:journals/csur/LuLHWC20, 7078994, pan2010survey, luo2022attention}. In this paper, we are mainly focused  on strengthening  \textbf{MMD}-based \textbf{DA} methods. We therefore overview the state-of-the-art \textbf{MMD-DA} techniques according to  the following two main research streams: 1) Shallow \textbf{MMD}-\textbf{DA}; 2) Deep \textbf{MMD}-\textbf{DA}, to better clarify the rationale of our proposed method. For comprehensiveness, we also discuss recent decision boundary optimization enforced \textbf{DA} to highlight the differences and the innovations of our proposed method.

\subsection{Shallow \textbf{MMD}-\textbf{DA}}
\label{subsect: Shallow DA}

Shallow \textbf{MMD}-\textbf{DA} aims to decrease the  \textit{domain shift} by minimizing a statistic measurement in a shared feature space across domains, \textit{i.e.}, \textbf{MMD}, within the traditional optimization-based shallow models. Typical approaches \cite{luo2020discriminative, long2013transfer, wang2020transfer} hybridize the merits of convex optimization, compressing sensing and manifold learning, \textit{etc}, to reduce the cross-domain divergence of feature representations while explicitly minimizing the \textbf{MMD} enforced distribution measurements, \textit{i.e.}, marginal distribution \cite{4967588} \cite{pan2011domain}, conditional distribution \cite{long2013transfer} or hybridized distrubution \cite{wang2020transfer}. In the search of such a domain shift reduced functional learning, these approaches can be further distinguished based on whether they incorporate some form of data discriminativeness or not.

\subsubsection{Nondiscriminative distribution alignment (\textbf{NDA})} \textbf{NDA} strategies propose to align the marginal and conditional distributions across the source and target domains in reducing the \textbf{MMD} induced distance measurement to explicitly shrink the cross-domain divergence of marginal data distributions  \cite{pan2011domain}, as well as the  marginal and conditional data distributions \cite{long2013transfer}. Lately, \textbf{ARTL} \cite{DBLP:journals/tkde/LongWDPY14} further hybridizes the distribution alignment and label propagation within a single unified optimization framework,  enabling the \textbf{NDA} techniques to also leverage data geometric knowledge. To avoid geometric structure distortion which could be present in the original feature space, \textbf{MEDA} \cite{wang2018visual} specifically learns the Grassmann manifold. The main drawback of \textbf{NDA} is that it does not explore data discriminativeness as induced by labeled data in the source domain, making it more difficult to search for a proper cross-domain classifier. This observation inspires the research communities for further exploration of the discriminative functional learning.

\subsubsection{Discriminative distribution alignment (\textbf{DDA})} \textbf{DDA} approaches improve \textbf{NDA} methods by incorporating data discriminativeness for the task-oriented model design.  \textbf{ILS} \cite{herath2017learning} learns a discriminative latent space using Mahalanobis metric and makes use of Riemannian optimization strategy to match statistical properties across different domains. \textbf{OBTL} \cite{karbalayghareh2018optimal} proposes bayesian transfer learning-based domain adaptation, which explicitly discusses the relatedness across different sub-domains.  \textbf{SCA} \cite{DBLP:journals/pami/GhifaryBKZ17} achieves discriminativeness in optimizing the interplay of both between and within-class scatters. \textbf{DGA-DA} \cite{luo2020discriminative} introduces a novel \textit{repulsive force} term to describe the data discriminativeness, which also optimizes the underlying data manifold structure when performing label inference.

As visualized in Fig.\ref{fig:2} (c),  \textbf{DDA} methods potentially reduce \textbf{Term.1} of Eq.(\ref{eq:bound}) in exploring the \textbf{MMD} induced data discriminativeness. However, such data discriminativeness is mainly focused on increasing the distances between the center points of different sub-domains and thus falls short in proactively handling the samples lying around the decision boundary. It therefore, is unable to generate the boundary aware domain adaptation as depicted in Fig.\ref{fig:2} (e) where data distributions are aligned across the two domains, inter-class distances are enlarged while intra-class samples are compacted.

\subsection{Deep learning-based \textbf{DA}}
\label{subsect: Deep DA}

Boosted by the success of the paradigm of deep learning (\textbf{DL}),  shallow \textbf{MMD-DA} approaches have been extended to DL-based ones. They can be distinguished based on whether they incorporate adversarial learning.


\subsubsection{MMD alignment based deep DA}

The principle of narrowing data distribution shift in shallow \textbf{MMD}-\textbf{DA} can be seamlessly embedded into deep models to formalize the \textit{\textbf{MMD} alignment-based deep \textbf{DA}}, thereby leveraging  highly discriminative deep features for further improved DA performance. Specifically, \textbf{DAN}\cite{long2015learning} reduces the marginal distribution divergence in incorporating the multi-kernel \textbf{MMD} loss on the fully connected layers of AlexNet. \textbf{JAN}\cite{DBLP:conf/icml/LongZ0J17} improves \textbf{DAN} by jointly decreasing the divergence of both the marginal and conditional distributions. \textbf{D-CORAL}\cite{sun2016deep} further introduces the second-order statistics into the AlexNet\cite{krizhevsky2012imagenet} framework for more effective \textbf{DA} strategy.

\subsubsection{Adversarial learning-based MMD-DA}
\label{Adversarial loss-based DA}
These methods make use of \textbf{GAN} \cite{goodfellow2014generative} and propose to align data distributions across domains in making sample features indistinguishable \textit{w.r.t} the domain labels through an adversarial loss on a domain classifier \cite{ganin2016domain,tzeng2017adversarial,pei2018multi}. \textbf{DANN} \cite{ganin2016domain} and \textbf{ADDA}\cite{tzeng2017adversarial} learn a  domain-invariant feature subspace in reducing the marginal distribution divergence. \textbf{MADA} \cite{pei2018multi} also makes use of multiple domain discriminators, thereby aligning conditional data distributions. Lately, \textbf{DSN}\cite{bousmalis2016domain} achieves domain-invariant representations in explicitly separating the similarities and dissimilarities in the source and target domains. Using multi-source domains, \textbf{MADAN} \cite{zhao2019multi} explores the multi-domain knowledge to fulfill \textbf{DA} tasks. \textbf{CyCADA} \cite{pmlr-v80-hoffman18a} addresses the distribution divergence using a bi-directional \textbf{GAN} based training framework.  \textbf{ATM} \cite{li2020maximum} specifically introduces maximum density divergence  (\textbf{MDD}), an original distance loss, into the adversarial learning framework to quantify distribution divergence for effective \textbf{DA}.

The main advantage of these \textbf{DL}-based \textbf{DA} methods is that they  shrink the across domain divergence of data distributions in deep features  and search at the same time an optimal classifier through a single unified end-to-end learning framework. Therefore, the optimized model naturally enjoys the merits similar to those of \textbf{META} learning \cite{wei2021toalign}. This suggests that the model can be further reinforced by harnessing different tasks so as to receive the best candidate hyper-parameters. However, when compared with traditional shallow \textbf{MMD}-\textbf{DA} approaches, they generally suffer from the noisy '\textit{batch learning}' \cite{goodfellow2016deep} strategy in contrast to '\textit{global modeling}' \cite{belkin2003laplacian} based optimization. Moreover, these approaches do not explicitly take into account decision boundaries  and ignore in particular to properly handle samples lying around decision boundaries for the decision boundary aware domain adaptation.

\subsection{Decision boundary optimization-based \textbf{DA}}
\label{subsect: Decision boundary optimization-based DA}

In line with max-margin confident prediction principle of classical semi-supervised learning \cite{roller2004max}, \textbf{D}ecision \textbf{B}oundary (\textbf{DB}) optimization aims to place class boundaries in low-density regions,  and shows its effectiveness in a number of machine learning tasks, \textit{i.e.}, active learning \cite{cho2022mcdal}, knowledge distillation \cite{heo2019knowledge}, domain adaptation, \textit{etc}. 

In domain adaptation, Lu \textit{et al.} \cite{lu2018embarrassingly} adopts the principle of linear discriminant analysis \cite{goodfellow2016deep} to optimize the \textbf{DB}, and demonstrate its ability to solve the cross-domain tasks even without explicit divergence reduction. \textbf{Asm-DA}\cite{DBLP:conf/icml/SaitoUH17}  propose asymmetric tri-training based \textbf{DA}, whereof  two auxiliary classifiers  trained on the source domain are encouraged to be highly different on the target domain, and show that their strategy optimizes the samples lying around the decision boundary for the \textbf{DB} optimized model training. Utilizing adversarial learning strategies, \textbf{Asm-DA} is further  improved by \textbf{MCD} \cite{DBLP:conf/cvpr/SaitoWUH18}, which dynamically optimizes the maximum classifier discrepancy regularized \textbf{DA}. Subsequently, \textbf{GPDA} \cite{kim2019unsupervised} hybridizes \textbf{MCD}’s principle and the Bayesian optimization framework to further simplify the learning paradigm. \textbf{BCMD}\cite{li2021bi} argues that the \textbf{DB} optimization without conditional distribution alignment may result to a deterioration of the representation discriminability, making it category agnostic. As a result, \textbf{BCMD} designs \textit{'classifier determinacy disparity metric'} to generate the conditional distribution alignment enforced \textbf{DB-DA}.

Inspired by the aforementioned research, we propose a novel decision boundary optimization enforced mechanism, namely, \textit{decision boundary aware graph}, which is composed of the '\textit{Compacting graph}' and 'Separation graph'. Specifically, the compacting graph aims to shrink each subdomain's divergence by properly regularizing the samples lying around decision boundaries, while the separation graph propagates the discriminativeness to further optimize  the cross-domain samples lying around decision boundaries for comprehensive decision boundary optimization. Therefore, the strengthened \textbf{DB-MMD} measurement discriminatively approaches the cross-domain adaptation and can seamlessly hybridize with the next round of classification for the decision boundary-aware \textbf{DA}.

\section{The propose method}
\label{The proposed method}

We define the notations and formalize the \textbf{DA} problem in Sect.\ref{Notations and Problem Statement}. Sect.\ref{Formulation} mathematically formulates the designed \textbf{DB-MMD} measurement and embeds the \textbf{DB-MMD} into a basic \textbf{DA} model, \textit{i.e.}, \textbf{CDDA} \cite{luo2020discriminative} for the decision boundary aware \textbf{DA}. Sect.\ref{Optimization}  presents the optimization process for solving the proposed \textbf{DA} method. Sect.\ref{Kernelization} extends the proposed \textbf{DA} method to non-linear problems through kernel mapping.

\subsection{Notations and Problem Statement}
\label{Notations and Problem Statement}

Matrices are represented in boldface uppercase letters. Vectors are represented in boldface lowercase letters. For matrix ${\bf{M}} = ({m_{ij}})$, its $i$-th row is denoted as ${{\bf{m}}^i}$, and its $j$-th column is denoted by ${{\bf{m}}_j}$.  We define the Frobenius norm ${\left\| . \right\|_F}$ norm as: ${\left\| {\bf{M}} \right\|_F} = \sqrt {\sum {_{i = 1}^n} \sum {_{j = 1}^l} m_{ij}^2} $. A domain $D$ is defined as an $\ell$-dimensional feature space $\chi$ and a marginal probability distribution $P(x)$, \textit{i.e.}, $\mathcal{D}=\{\chi,P(x)\}$ with $x\in \chi$.  Given a specific domain $D$, a  task $T$ is composed of a C-cardinality label set $\mathcal{Y}$  and a classifier $f(x)$,\textit{ i.e.}, $T = \{\mathcal{Y},f(x)\}$, where $f({x}) = \mathcal{Q}( y |x)$ can be interpreted as the class conditional probability distribution for each input sample $x$.

In unsupervised domain adaptation, we are given a source domain $\mathcal{D_S}=\{x_{i}^{s},y_{i}^{s}\}_{i=1}^{n_s}$ with $n_s$ labeled samples ${{\bf{X}}_{\cal S}} = [x_1^s...x_{{n_s}}^s]$, which are associated with their class labels ${{\bf{Y}}_S} = {\{ {y_1},...,{y_{{n_s}}}\} ^T} \in {{\bf{\mathbb{R}}}^{{n_s} \times C}}$, and an unlabeled target domain $\mathcal{D_T}=\{x_{j}^{t}\}_{j=1}^{n_t}$ with $n_t$  unlabeled samples ${{\bf{X}}_{\cal T}} = [x_1^t...x_{{n_t}}^t]$, whose labels  ${{\bf{Y}}_T} = {\{ {y_{{n_s} + 1}},...,{y_{{n_s} + {n_t}}}\} ^T} \in {{\bf{\mathbb{R}}}^{{n_t} \times C}}$ are unknown. Here, ${y_i} \in {{\bf{\mathbb{R}}}^c}(1 \le i \le {n_s} + {n_t})$ is a one-vs-all label hot vector in which $y_i^j = 1$ if ${x_i}$ belongs to the $j$-th class, and $0$ otherwise. We  define the data matrix ${\bf{X}} = [{{\bf{X}}_S},{{\bf{X}}_T}] \in {R^{l*n}}$ ($l = feature\;\dim ension$; $n = {n_s} + {n_t}$ ) in packing both the source and target data. The source domain $\mathcal{D_S}$ and target domain $\mathcal{D_T}$ are assumed to be different, \textit{i.e.},  $\mathcal{\chi}_S=\mathcal{{\chi}_T}$, $\mathcal{Y_S}=\mathcal{Y_T}$, $\mathcal{P}(\mathcal{\chi_S}) \neq \mathcal{P}(\mathcal{\chi_T})$, $\mathcal{Q}(\mathcal{Y_S}|\mathcal{\chi_{S}}) \neq \mathcal{Q}(\mathcal{Y_T}|\mathcal{\chi_{T}})$. We also define the notion of \textit{sub-domain}, \textit{i.e.}, class,  denoted as ${\cal D}_{\cal S}^{(c)}$, representing the set of samples in ${{\cal D}_{\cal S}}$ with the class label $c$. It is noteworthy that, the definition of sub-domains in the target domain, namely ${\cal D}_{\cal T}^{(c)}$,  requires a base classifier,\textit{ e.g.}, Nearest Neighbor (NN),  to attribute  pseudo labels for samples in ${{\cal D}_{\cal T}}$.

\textbf{MMD}: The  maximum mean discrepancy (MMD)  is an effective non-parametric distance-measure  that compares the distributions of two sets of data by mapping the data into Reproducing Kernel Hilbert Space\cite{borgwardt2006integrating} (RKHS). Given two distributions $\mathcal{P}$ and $\mathcal{Q}$, the MMD between $\mathcal{P}$ and $\mathcal{Q}$ is defined as:

	\begin{equation}\label{eq:MMD}
	Dist({\mathcal P},{\mathcal Q}) = {(\int_{\mathcal P} {\phi ({p_i})} {\rm{ }}{d_{{p_i}}} - \int_{\mathcal Q} {\phi ({q_i})} {\rm{ }}{d_{{q_i}}})_{\mathcal H}}
\end{equation}

where ${\mathcal P}=\{ p_1, \ldots, p_{n_1}, \ldots \}$ and ${\mathcal Q} = \{ q_1, \ldots, q_{n_2}, \ldots \}$ are two random variable sets from distributions $\mathcal{P}$ and $\mathcal{Q}$, respectively, and $\mathcal{H}$ is a universal RKHS with the reproducing kernel mapping $\phi$: $f(x) = \langle \phi(x), f \rangle$, $\phi: \mathcal{X} \to \mathcal{H}$. In real practice, the \textbf{MMD} is estimated on finite samples:

	\begin{equation}\label{eq:MMD1}
	Dist({\mathcal P},{\mathcal Q}) = \parallel \frac{1}{n_1} \sum^{n_1}_{i=1} \phi(p_i) - \frac{1}{n_2} \sum^{n_2}_{i=1} \phi(q_i) \parallel_{\mathcal{H}}
\end{equation}

where ${p_i}$ and ${p_i}$ are independent random samples drawn from the distributions $\mathcal{P}$ and $\mathcal{Q}$ respectively.

 \textbf{Affinity matrix}: Affinity matrix ${\bf{W}}$ is proposed to capture the relative distance of each sample across the entire dataset. Formally, we denote the pair-wise affinity matrix as:

	\begin{equation}\label{eq:2}
	{{\mathbf{W}}_{ij = }}\left\{ {\begin{array}{*{20}{c}}
  {sim({x_i},{x_j}),\;\;{x_i} \in {N_p}({x_j})\;or\;{x_j} \in {N_p}({x_i})} \\ 
  {0,\;\;\;\;\;\;\;\;\;\;\;\;\;\;\;otherwise\;\;\;\;\;\;\;\;\;\;\;\;\;\;\;\;\;\;\;\;\;\;\;} 
\end{array}} \right.
\end{equation}

where  ${\bf{W}} = {[{w_{ij}}]_{({n_s} + {n_t}) \times ({n_s} + {n_t})}}$ is a symmetric matrix \cite{NIPS2001_2092}, with   ${w_{ij}}$ giving the affinity between two data samples $i$ and $j$, and is defined as ${sim({x_i},{x_j})} = \exp ( - \frac{{{{\left\| {{\bf{x}_i} - {\bf{x}_j}} \right\|}^2}}}{{2{\sigma ^2}}})$ if $i \ne j$ and ${w_{ii}} = 0$. Therefore, the larger distance between different samples ${dist({x_i},{x_j})}$ leading low similarity value ${sim({x_i},{x_j})}$, and vice versa.




\subsection{\textbf{DB-MMD} Formulation}
\label{Formulation}

As highlighted in Fig.\ref{fig:2}, the key mathematical formulations of the \textbf{DB-MMD} measurement are as follows:

\begin{itemize}
	\item Sect.\ref{subsection:Matching Marginal and Conditional Distributions} (\textbf{Cross Domain/Sub-domain Distribution Alignment}): As illustrated in Fig.\ref{fig:2}.(b),  \textbf{DB-MMD} begins by aligning the cross-domain marginal/conditional distribution to minimize  \textbf{Term.2} of the hypothesis error bound (Eq.(\ref{eq:bound})).

	\item Sect.\ref{Repulsing interclass data for discriminative DA} (\textbf{Cross Sub-domains Separation}): 
	As in Fig.\ref{fig:2}.(c), the \textit{repulsive force} (\textbf{RF}) term is proposed to drag away the cross sub-domains with different labels, thereby  facilitating classification on the source domain, and thus minimizing  \textbf{Term.1} in Eq.(\ref{eq:bound}).

	\item Sect.\ref{Alignment Graph} (\textbf{Decision Boundary aware Graph}):  As depicted in Fig.\ref{fig:2}.(d), a \textbf{compacting graph} and a \textbf{separation graph} are designed to further improve Sect.\ref{subsection:Matching Marginal and Conditional Distributions} and Sect.\ref{Repulsing interclass data for discriminative DA}, respectively, to compact intra-class samples  while separating inter-class samples across the domains,  thereby optimizing the decision boundary for the next round of classification and minimizing \textbf{Term.3} in Eq.(\ref{eq:bound}).
	
    	\item Sect.\ref{Final model} (\textbf{Decision Boundary Optimization-Informed Domain Adaptation}): The proposed \textbf{DB-MMD} is formulated by hybridizing the properties as presented in Sect.\ref{subsection:Matching Marginal and Conditional Distributions}, Sect.\ref{Repulsing interclass data for discriminative DA} and Sect.\ref{Alignment Graph}. The proposed \textbf{DB-MMD} is then embedded into a baseline \textbf{DA} model, \textit{i.e.}, \textbf{CDDA},  to generate a novel decision boundary aware \textbf{DA}.
	
 	\end{itemize}

\subsubsection{Marginal and Conditional Distribution Alignment}
\label{subsection:Matching Marginal and Conditional Distributions}

Following the popular \textbf{JDA} \cite{long2013transfer}, the  cross-domain divergence can be formulated as:
	
		\begin{equation}\label{eq:term2}
	\left\{ \begin{gathered}
  {\mathbf{1}}{\mathbf{.}}{\text{ }}\mathcal{P}({\mathcal{X}_\mathcal{S}}) \ne \mathcal{P}({\mathcal{X}_\mathcal{T}}),\mathcal{Q}({\mathcal{Y}_\mathcal{S}}|{\mathcal{X}_\mathcal{S}}) \ne \mathcal{Q}({\mathcal{Y}_\mathcal{T}}|{\mathcal{X}_\mathcal{T}}) \hfill \\
  {\mathbf{2}}{\mathbf{.}}{\text{ }}{\mathcal{X}_S} = {\mathcal{X}_\mathcal{T}},{\mathcal{Y}_\mathcal{S}} = {\mathcal{Y}_\mathcal{T}} \hfill \\ 
\end{gathered}  \right.
\end{equation}

Eq.(\ref{eq:term2}.2) states that the source domain ($\mathcal{D_S}$) and the target domain ($\mathcal{D_T}$) share a same feature (${\cal X}$) and label space (${\cal Y}$), while Eq.(\ref{eq:term2}.1) says that the two domains are different from each other in terms of the marginal/conditional probability distribution. In this research, \textbf{MMD} in \textbf{RKHS} is used to measure the distances between the expectations of the source domain/sub-domain and target domain/sub-domain and to quantify the existing domain divergence. Specifically, \textbf{1)} the empirical distance of the source and target domains is defined as $Dis{t^{m}}$; and \textbf{2)} the conditional distance $Dis{t^{c}}$ is defined as the sum of the empirical distances between sub-domains in ${{\cal D}_{\cal S}}$ and ${{\cal D}_{\cal T}}$ with a same label.

\begin{equation}\label{eq:3.1.2}
	\left\{ \begin{gathered}
  {\mathbf{1}}{\mathbf{.}}{\text{ }}Dis{t^m} = {(\int_{{\mathcal{D}_\mathcal{S}}} {\phi ({x_i})} {d_{{x_i}}} - \int_{{\mathcal{D}_\mathcal{T}}} {\phi ({x_j})} {d_{{x_j}}})_\mathcal{H}} \hfill \\
  {\mathbf{2}}{\mathbf{.}}{\text{ }}Dis{t^c} = \sum\limits_{c = 1...C} {} {(\int_{\mathcal{D}_\mathcal{S}^{(c)}} {\phi ({x_i})} {d_{{x_i}}} - \int_{\mathcal{D}_\mathcal{T}^{(c)}} {\phi ({x_j})} {d_{{x_j}}})_\mathcal{H}} \hfill \\ 
\end{gathered}  \right.
\end{equation}

	where $C$ is the number of classes, $\mathcal{D_S}^{(c)} = \{ {\bf{x}_i}:{\bf{x}_i} \in \mathcal{D_S} \wedge y({\bf{x}_i}) = c\} $ represents the ${c^{th}}$ sub-domain in the source domain,  $n_s^{(c)} = {\left\| {\mathcal{D_S}^{(c)}} \right\|_0}$  the number of samples in the ${c^{th}}$ {source} sub-domain. $\mathcal{D_T}^{(c)}$ and $n_t^{(c)}$ are defined similarly for the target domain but using pseudo-labels. As a result, in Eq.(\ref{eq:3.1.2}), the divergence between the cross-domain marginal distributions and the one between conditional distributions are reduced in minimizing ${Dis{t^{m}}}$ and ${Dis{t^{c}}}$, respectively.

\begin{itemize}

	\item  \textbf{Implementation:}
\end{itemize}

	In real-life applications, we have a finite number of samples,  Eq.(\ref{eq:3.1.2}) is reformulated as Eq.(\ref{eq:JDA}), where $Dis{t_{Clo}}$ is defined as the sum of $Dis{t^{m}}$ and $Dis{t^{c}}$.

\begin{equation}\label{eq:JDA}
\begin{array}{l}
		Dis{t_{Clo}} = Dis{t^m}({D_S},{D_T}) + Dis{t^c}\sum\limits_{c = 1}^C {({D_S}^c,{D_T}^c)} \\
		\;\;\;\;\;\;\;\;\; = {\left\| {\frac{1}{{{n_s}}}\sum\limits_{i = 1}^{{n_s}} {{{\bf{A}}^T}{x_i} - } \frac{1}{{{n_t}}}\sum\limits_{j = {n_s} + 1}^{{n_s} + {n_t}} {{{\bf{A}}^T}{x_j}} } \right\|^2} + {\left\| {\frac{1}{{n_s^{(c)}}}\sum\limits_{{x_i} \in {D_S}^{(c)}} {{{\bf{A}}^T}{x_i}}  - \frac{1}{{n_t^{(c)}}}\sum\limits_{{x_j} \in {D_T}^{(c)}} {{{\bf{A}}^T}{x_j}} } \right\|^2}\\
		\;\;\;\;\;\;\;\;\; = tr({{\bf{A}}^T}{\bf{X}}({{\bf{M}}_{\bf{0}}} + \sum\limits_{c = 1}^{c = C} {{{\bf{M}}_c}} ){{\bf{X}}^{\bf{T}}}{\bf{A}})
		\end{array}
\end{equation}

\begin{itemize}

	\item {$Dis{t^{m}}({{\cal D}_{\cal S}},{{\cal D}_{\cal T}})$}: where ${{\bf{M}}_0}$ is the MMD matrix  between ${{\cal D}_{\cal S}}$ and ${{\cal D}_{\cal T}}$ with ${{{({{\bf{M}}_0})}_{ij}} = \frac{1}{{{n_s}{n_s}}}}$ if $({{\bf{x}_i},{\bf{x}_j} \in {D_S}})$, ${{{({{\bf{M}}_0})}_{ij}} = \frac{1}{{{n_t}{n_t}}}}$ if $({{\bf{x}_i},{\bf{x}_j} \in {D_T}})$ and ${({{\bf{M}}_0})_{ij}} = \frac{{ - 1}}{{{n_s}{n_t}}}$ otherwise.  Thus, the difference between the marginal distributions $\mathcal{P}(\mathcal{X_S})$ and $\mathcal{P}(\mathcal{X_T})$ is reduced when minimizing {$Dis{t^{m}}({{\cal D}_{\cal S}},{{\cal D}_{\cal T}})$}.

   \item {$Dis{t^{c}}({{\cal D}_{\cal S}},{{\cal D}_{\cal T}})$}:	 where $\bf M_c$ denotes the MMD matrix between the sub-domains with labels $c$ in ${{\cal D}_{\cal S}}$ and ${{\cal D}_{\cal T}}$ with ${{{({{\bf{M}}_c})}_{ij}} = \frac{1}{{n_s^{(c)}n_s^{(c)}}}}$ if $({{\bf{x}_i},{\bf{x}_j} \in {D_S}^{(c)}})$, ${{{({{\bf{M}}_c})}_{ij}} = \frac{1}{{n_t^{(c)}n_t^{(c)}}}}$ if $({{\bf{x}_i},{\bf{x}_j} \in {D_T}^{(c)}})$, ${{{({{\bf{M}}_c})}_{ij}} = \frac{{ - 1}}{{n_s^{(c)}n_t^{(c)}}}}$ if $({{\bf{x}_i} \in {D_S}^{(c)},{\bf{x}_j} \in {D_T}^{(c)}\;or\;\;{\bf{x}_i} \in {D_T}^{(c)},{\bf{x}_j} \in {D_S}^{(c)}})$ and ${{{({{\bf{M}}_c})}_{ij}} = 0}$ otherwise. Consequently, the mismatch of conditional distributions between ${{D_{\cal S}}^c}$ and ${{D_{\cal T}}^c}$ is reduced by minimizing ${Dis{t^{c}}}$. 
   
\end{itemize}


\subsubsection{Cross Sub-domains Separation}
\label{Repulsing interclass data for discriminative DA}

In addition to the minimization of \textbf{Term.2} of Eq.(\ref{eq:bound}) as in Sect.\ref{subsection:Matching Marginal and Conditional Distributions}, we aim in Sect.\ref{Repulsing interclass data for discriminative DA} to decrease \textbf{Term.1} of Eq.(\ref{eq:bound}), namely, \textit{classification error on the source domain}:

\begin{equation}\label{eq:error}
{e_\mathcal{S}}(h) = \sum\limits_{i = 1,{x_i} \in {\mathcal{D}_\mathcal{S}}}^{{n_s}} {\left\| {{\mathcal{Y}_{{x_i} \in \mathcal{D}_\mathcal{S}^{(c)}}} \ne c} \right\|}
\end{equation}		
	
 where 	${e_\mathcal{S}}(h)$ denotes all the source domain samples with incorrect labels. Minimization of  Eq.(\ref{eq:error}) can be approached by optimizing Eq.(\ref{eq:3.2.1}):


\begin{equation}\label{eq:3.2.1}
	\mathop {\min }\limits_{c \in \{ 1...C\} } \sum\limits_{,i = 1}^{{n_s}} {\mathcal{Q}(\mathcal{Y} \ne c|{x_i} \in \mathcal{D}_\mathcal{S}^{(c)})} {\text{ }}.
\end{equation}

A straightforward solution to  Eq.(\ref{eq:3.2.1}) is to explicitly increase the distribution divergence between each sub-domain ${\mathcal{D}_{\mathcal{S}}}^{(c)}$ and all the other source sub-domains ${\mathcal{D}_{\cal \mathcal{S}}}^{(r);\;r \in \{ \{ 1...C\}  - \{ c\} \} }$, making all the different sub-domains well separated. For this purpose, 	in Eq.(\ref{eq:3.2.2}), we design the \textit{repulsive force}(\textbf{RF}) term ($Dis{t^{re}}$) to quantify the overall divergence between ${\mathcal{D}_{\mathcal{S}}}^{(c)}$ and ${\mathcal{D}_{\mathcal{S}}}^{(r)}$:

	\begin{equation}\label{eq:3.2.2}
	Dis{t^{re}} = \sum\limits_{c = 1...C} {} {(\int_{\mathcal{D}_\mathcal{S}^c} {\phi ({x_i})} {d_{{x_i}}} - \int_{\mathcal{D}_\mathcal{S}^{r \in \{ \{ 1...C\}  - \{ c\} \} }} {\phi ({x_j})} {d_{{x_j}}})_\mathcal{H}}.
\end{equation}	
	
In Eq.(\ref{eq:3.2.2}), the \textit{repulsive force}(\textbf{RF}) term ($Dis{t^{re}}$) is designed to enforce the discriminative functional learning on ${\mathcal{D}_\mathcal{S}}$. We further improve Eq.(\ref{eq:3.2.2}) and formulate Eq.(\ref{eq:3.2.3}) to comprehensively propagate the discriminativeness across both domains $({\mathcal{D}_\mathcal{S}},{\mathcal{D}_\mathcal{T}})$.

\begin{equation}\label{eq:3.2.3}
	\left\{ \begin{gathered}
  {\mathbf{1}}{\mathbf{.}}{\text{ }}Dist_{\mathcal{S} \to \mathcal{T}}^{re} = \sum\limits_{c = 1...C} {} {(\int_{\mathcal{D}_\mathcal{S}^c} {\phi ({x_i})} {d_{{x_i}}} - \int_{\mathcal{D}_\mathcal{T}^{r \in \{ \{ 1...C\}  - \{ c\} \} }} {\phi ({x_j})} {d_{{x_j}}})_\mathcal{H}} \hfill \\
  {\mathbf{2}}{\mathbf{.}}{\text{ }}Dist_{\mathcal{T} \to \mathcal{S}}^{re} = \sum\limits_{c = 1...C} {} {(\int_{\mathcal{D}_\mathcal{T}^c} {\phi ({x_j})} {d_{{x_j}}} - \int_{\mathcal{D}_\mathcal{S}^{^{r \in \{ \{ 1...C\}  - \{ c\} \} }}} {\phi ({x_i})} {d_{{x_i}}})_\mathcal{H}} \hfill \\ 
\end{gathered}  \right.
\end{equation}

where ${{\cal S} \to {\cal T}}$ and ${{\cal T} \to {\cal S}}$  index the distances computed from '${D_{\cal S}}$ to ${D_{\cal T}}$' and  '${D_{\cal T}}$ to ${D_{\cal S}}$', respectively. As can be seen,  $Dist_{{\cal S} \to {\cal T}}^{re}$ and $Dist_{{\cal T} \to {\cal S}}^{re}$ are defined in a similar way as $Dis{t^{re}}$ in Eq.(\ref{eq:3.2.2}).  The rationale and merits of reformulating Eq.(\ref{eq:3.2.2}) as  Eq.(\ref{eq:3.2.3}) are summarized as follows:

\textbf{Rationale}: The discriminativeness characterized by Eq.(\ref{eq:3.2.2}) can be achieved by Eq.(\ref{eq:3.2.3}.1) and Eq.(\ref{eq:3.2.3}.2), respectively.

\textit{\textbf{Proof}}: In Sect.\ref{subsection:Matching Marginal and Conditional Distributions}, the conditional distribution divergence between the cross sub-domains ${(\mathcal{D}_\mathcal{S}^c,\mathcal{D}_\mathcal{T}^c)_{c \in \{ 1...C\} }}$ is reduced by minimizing Eq.(\ref{eq:3.1.2}.2), then all the cross sub-domains (${(\mathcal{D}_\mathcal{S}^c,\mathcal{D}_\mathcal{T}^c)_{c \in \{ 1...C\} }}$) are well aligned, leading to the similar kernel mean embeddings between the cross sub-domain pairs:

\begin{equation}\label{eq:3.2.5}
	\mathop \forall \limits_{c = 1...C} {(\int_{D_{\cal S}^{(c)}} {\phi ({x_j})} {d_{{x_j}}} = \int_{{\cal D}_T^{(c)}} {\phi ({x_j})} {d_{{x_j}}})_{\cal H}}.
\end{equation}	

Following Eq.(\ref{eq:3.2.5}), we can simply replace the $\int_{\mathcal{D}_\mathcal{T}^c} {\phi ({x_j})} {d_{{x_j}}}$ term in Eq.(\ref{eq:3.2.3}.2) by $\int_{\mathcal{D}_\mathcal{S}^c} {\phi ({x_j})} {d_{{x_j}}}$ , then Eq.(\ref{eq:3.2.3}.2) is naturally reformulated as Eq.(\ref{eq:3.2.2}). Similarly, Eq.(\ref{eq:3.2.3}.1) can be reformulated as Eq.(\ref{eq:3.2.2}).

\textit{\textbf{Merits}}:  Eq.(\ref{eq:3.2.3}) improves Eq.(\ref{eq:3.2.2}) by dynamically propagating data discriminativeness  across both domains, thus beyond a single domain, within the searched feature space. In such a way, the cross-domain labeling functions across domains are unified, thereby enabling  a cycle-consistent  \cite{zhu2017unpaired} like learning enforced model training.

\begin{itemize}

	\item  \textbf{Implementation:}
	\end{itemize}

	In practice, Eq.(\ref{eq:3.2.3}) is reformulated as Eq.(\ref{eq:CDDAnew}) by using a finite number of samples:
	
\begin{equation}\label{eq:CDDAnew}
	\begin{array}{l}
		Dist_{S \to T}^{re} + Dist_{T \to S}^{re} \\
		 =  Dis{t^c}\sum\limits_{c = 1}^C {({D_S}^c,{D_T}^{r \in \{ \{ 1...C\}  - \{ c\} \} })} + Dis{t^c}\sum\limits_{c = 1}^C {({D_T}^c,{D_S}^{r \in \{ \{ 1...C\}  - \{ c\} \} })} \\
		 = \sum\limits_{c = 1}^C {tr({{\bf{A}}^T}{\bf{X}}({{\bf{M}}_{S \to T}} + {{\bf{M}}_{T \to S}}){{\bf{X}}^{\bf{T}}}{\bf{A}})} 
		\end{array}
\end{equation}

\begin{itemize}
	\item {${{\bf{M}}_{{\cal S} \to {\cal T}}}$ is defined as: ${{{({{\bf{M}}_{{\bf{S}} \to {\bf{T}}}})}_{ij}} = \frac{1}{{n_s^{(c)}n_s^{(c)}}}}$ if $({{\bf{x}_i},{\bf{x}_j} \in {D_S}^{(c)}})$, ${  \frac{1}{{n_t^{(r)}n_t^{(r)}}}}$ if $({{\bf{x}_i},{\bf{x}_j} \in {D_T}^{(r)}})$, ${  \frac{{ - 1}}{{n_s^{(c)}n_t^{(r)}}}}$ if $({{\bf{x}_i} \in {D_S}^{(c)},{\bf{x}_j} \in {D_T}^{(r)}\;or\;{\bf{x}_i} \in {D_T}^{(r)},{\bf{x}_j} \in {D_S}^{(c)}})$ and ${  0}$ otherwise.}
	
	\item {${{\bf{M}}_{{\cal T} \to {\cal S}}}$ is defined as: ${{{({{\bf{M}}_{{\bf{T}} \to {\bf{S}}}})}_{ij}} = \frac{1}{{n_t^{(c)}n_t^{(c)}}}}$ if ${({\bf{x}_i},{\bf{x}_j} \in {D_T}^{(c)})}$, ${  \frac{1}{{n_s^{(r)}n_s^{(r)}}}}$ if ${({\bf{x}_i},{\bf{x}_j} \in {D_S}^{(r)})}$, ${  \frac{{ - 1}}{{n_t^{(c)}n_s^{(r)}}}}$ if ${({\bf{x}_i} \in {D_T}^{(c)},{\bf{x}_j} \in {D_S}^{(r)}\;or\;{\bf{x}_i} \in {D_S}^{(r)},{\bf{x}_j} \in {D_T}^{(c)})}$ and ${  0}$ otherwise.}
	\end{itemize}

Therefore, maximizing Eq.(\ref{eq:CDDAnew}) facilitates a discriminative \textbf{DA}, thereby reducing the source domain error (\textbf{Term.1} of Eq.(\ref{eq:bound})).

\subsubsection{\textit{Decision Boundary aware Graph} }
\label{Alignment Graph}

\begin{figure}[h!]
	\centering
	\includegraphics[width=0.65\linewidth]{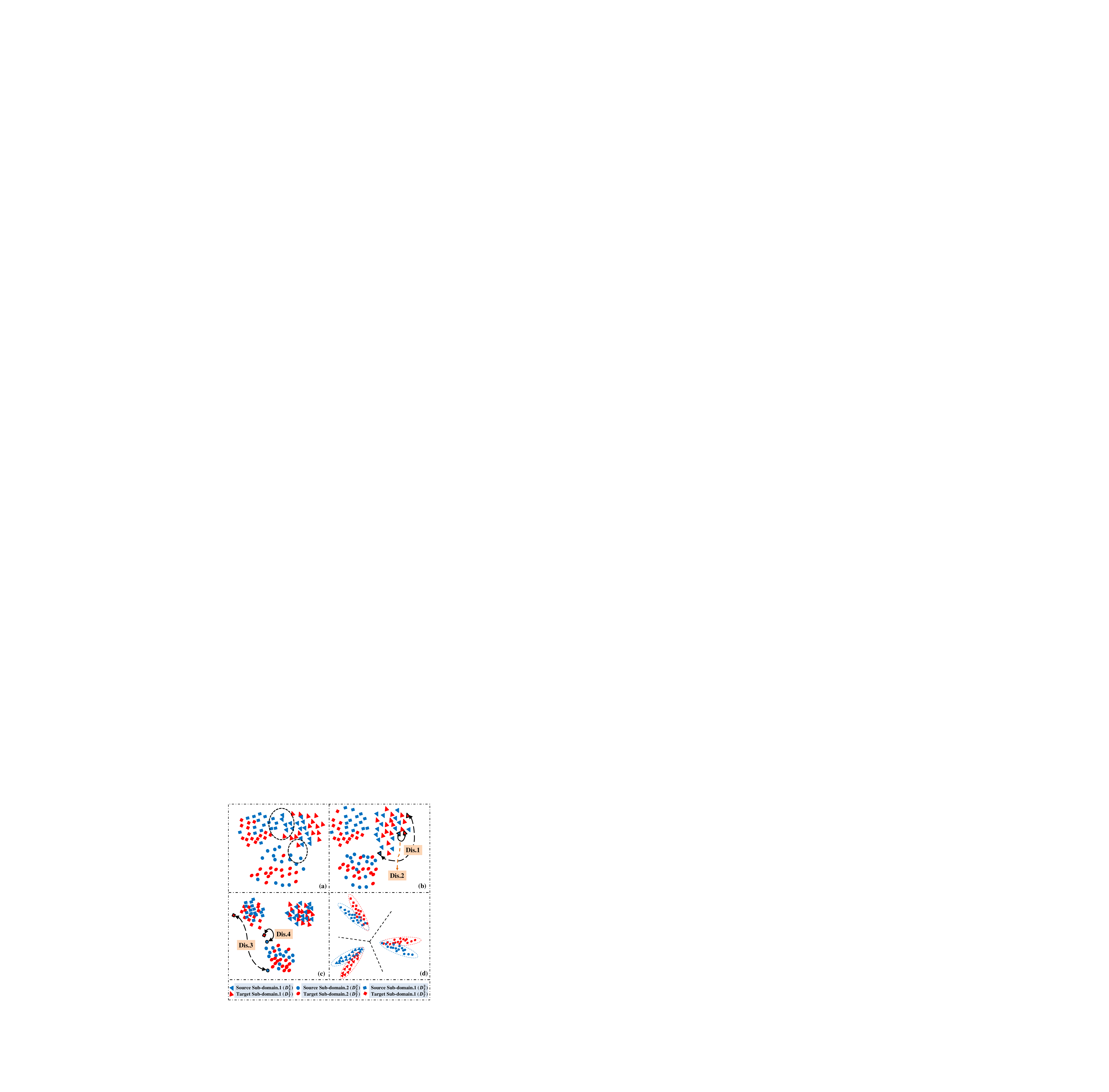}

	\caption {In Fig.\ref{fig:graph1}.(a), \textbf{DA} explores the effectiveness of  \textbf{distribution alignment} to drag close the domains and the sub-domains, while ignoring to optimize the samples lying around decision boundaries (Fig.\ref{fig:graph1}.(b,c)) for generating a \textbf{decision boundary optimization} guaranteed functional learning as illustrated in Fig.\ref{fig:graph1}.(d).}
		\label{fig:graph1}
\end{figure}

As discussed in Sect.(\ref{subsection:Matching Marginal and Conditional Distributions}$\&$\ref{Repulsing interclass data for discriminative DA}), although \textbf{Distribution Alignment} has shown its effectiveness in a number of cross-domain \textbf{DA} tasks \cite{lu2018embarrassingly,long2013transfer,DBLP:journals/csur/LuLHWC20,pan2010survey,7078994}, it faces difficulties of finding an optimized classification boundary for the decision boundary aware \textbf{DA}  when samples lying at boundaries from different sub-domains are mixed up. Indeed, as highlighted in Fig.\ref{fig:graph1}, plain vanilla \textbf{MMD} measurement merely cares about the overall statistical properties and treats each sample with equal weight, thereby making it unable to enforce  sample dependent yet appropriate constraint.

\begin{itemize}
	\item As can be seen in Fig.\ref{fig:graph1}.(b) , the two dashed lines, \textit{i.e.}, \textbf{Dis.1} and \textbf{Dis.2}, denote two different distances of two cross domain sample pairs of the same class, with \textbf{Dis.1} denoting a distance of two boundary samples. While   minimization of Eq.(\ref{eq:JDA}) in Sect.(\ref{subsection:Matching Marginal and Conditional Distributions}) drags closer the domain/sub-domain for distribution alignment, a better training process should pay more attention to reduce \textbf{Dis.1} rather than \textbf{Dis.2} for a decision boundary aware \textbf{DA} model. Nevertheless, using Eq.(\ref{eq:JDA}) enforced \textbf{MMD} measurement, the different cross sub-domain samples are assigned with similar weights, \textit{i.e.}, ${\frac{1}{{n_s^{(c)}}}}$ and ${\frac{1}{{n_t^{(c)}}}}$, respectively. As a result, \textbf{Dis.1} and \textbf{Dis.2} are regularized using a same fixed weight $(\frac{1}{{n_s^{(1)}}} * \frac{1}{{n_t^{(1)}}})$.

	\item Similarly, in Fig.\ref{fig:graph1}.(c), \textbf{Dis.3} and \textbf{Dis.4} are two distances of two inter-class sample pairs with \textbf{Dis.4} representing the distance of two close inter-class samples. While maximization of Eq.(\ref{eq:CDDAnew}) as introduced in Sect.(\ref{Repulsing interclass data for discriminative DA}) enables the cross sub-domains separation, \textbf{Dis.3} and \textbf{Dis.4} receive a same regularization weight (${(\frac{1}{{n_s^{(2)}}}*\frac{1}{{n_t^{(3)}}})}$), it  neglects to provide more attention to increase \textbf{Dis.4} rather than \textbf{Dis.3} for decision boundary optimization. 
	
	\item In summary, \textbf{Distribution Alignment} as elaborated in Sect.(\ref{subsection:Matching Marginal and Conditional Distributions}$\&$\ref{Repulsing interclass data for discriminative DA}) ignores the decision boundary awareness and thereby makes it hard to specifically serve for the \textbf{Classifier Optimization}. 
	
\end{itemize}

To fight the aforementioned weaknesses, we introduce here the decision boundary aware graph. It consists of a \textbf{compacting} and a \textbf{separation graph}, to make Eq.(\ref{eq:JDA}) and Eq.(\ref{eq:CDDAnew}) aware of decision boundaries.


	\textbf{1). Compacting Graph}: To improve Eq.(\ref{eq:JDA}), our aim is to drag closer far away separated sample pairs, \textit{e.g.}, \textbf{Dis.1} in Fig.\ref{fig:graph1}.(b), with more attention. Specifically, let  ($({({x_i})_{ \in {D_S}^c}},{({x_j})_{ \in {D_T}^c}})$) designate an intra-class sample pair across domain, \textit{i.e.}, $({x_i},{x_j})$ come from different domains but have a same label. We define the affinity value ${W_{ij}} = \exp ( - \frac{{{{\left\| {{{\bf{x}}_{\bf{i}}} - {{\bf{x}}_{\bf{j}}}} \right\|}^2}}}{{2{\sigma ^2}}})$ as in Eq.(\ref{eq:2}). As a result, when $({x_i},{x_j})$ is a pair of samples far away each other,  $({x_i},{x_j})$ receive small affinity value. We formulate Eq.(\ref{eq:4})

	\begin{equation}\label{eq:4}
	\mathop {\max \sum\limits_{c = 1}^C {(\frac{{(x_i^T{x_j})}}{{{W_{ij}}}})} }\limits_{s.t.{{({x_i})}_{ \in {D_S}^c}},{{({x_j})}_{ \in {D_T}^c}}}  \Rightarrow \mathop {\min \sum\limits_{c = 1}^C {( - \frac{{(x_i^T{x_j})}}{{{W_{ij}}}})} }\limits_{s.t.{{({x_i})}_{ \in {D_S}^c}},{{({x_j})}_{ \in {D_T}^c}}}.
\end{equation}

On the left-hand of Eq.(\ref{eq:4}),  sample pairs with large relative distance
, \textit{e.g.}, ${\mathbf{Dist}}{\mathbf{.}}{{\mathbf{1}}_{({x_i},{x_j})}}$, is naturally enforced with large weight=($1/{W_{ij}}$) \cite{NIPS2001_2092}. Thus, maximizing the left-hand of Eq.(\ref{eq:4}) provides more strengths to increase ${x_i^T{x_j}}$ so as to shrink the relative distance of ${\mathbf{Dist}}{\mathbf{.}}{{\mathbf{1}}_{({x_i},{x_j})}}$, thereby compacting intra-class samples and optimizing the decision boundaries. In the final model, the maximization on the left-hand of  Eq.(\ref{eq:4}) is reformulated as a problem of minimization as on the right-hand of Eq.(\ref{eq:4}). Through simple mathematical operations, Eq.(\ref{eq:4}) can be reformulated as:

	\begin{equation}\label{eq:5}
	\begin{gathered}
  \min ({\mathbf{X}}({{\mathbf{G}}_{{\mathbf{CG}}}}){{\mathbf{X}}^{\mathbf{T}}}) \hfill \\
  s.t.\;\;{{\mathbf{G}}_{{\mathbf{CG}}}} =  - (1/{\mathbf{W}}).*{\mathbf{MAS}}{{\mathbf{K}}_{({n_s} + {n_t},{n_s} + {n_t})}}\; \hfill \\
  \;\;\;\;\;\;\left\{ \begin{gathered}
  {\mathbf{MAS}}{{\mathbf{K}}_{ij}} = 1,\;\;if\;{({{\mathbf{M}}_{c \in \{ 1...C\} }})_{ij}} = \frac{{ - 1}}{{n_s^{(c)}n_t^{(c)}}} \hfill \\
  {\mathbf{MAS}}{{\mathbf{K}}_{ij}} = 0,\;\;otherwise \hfill \\ 
\end{gathered}  \right. \hfill \\ 
\end{gathered} 
\end{equation}

where ${{\mathbf{M}}_{c \in \{ 1...C\} }}$ is similarly defined as in Eq.(\ref{eq:JDA}). In Eq.(\ref{eq:5}), ${{\bf{G}}_{{\bf{CG}}}}$ is the \textbf{Compacting Graph}, which hybridizes with Eq.(\ref{eq:JDA}) to further nowrrow the within-domain divergence for a decision boundary aware cross domain distribution alignment.



	\textbf{2). Separation Graph}:  In order to increase \textbf{Dis.4} rather than \textbf{Dis.3} as highlighted in Fig.\ref{fig:graph1}.(c) through Eq.(\ref{eq:CDDAnew}), we propose a  \textbf{Separation Graph} to pay more attention on these inter-class cross domain sample pairs ($({({x_i})_{ \in {D_S}^c}},{({x_j})_{ \notin {D_T}^c}})$) where $({x_i},{x_j})$  belong to different domains and have different labels. Specifically, using the affinity  ${W_{ij}} = \exp ( - \frac{{{{\left\| {{{\bf{x}}_{\bf{i}}} - {{\bf{x}}_{\bf{j}}}} \right\|}^2}}}{{2{\sigma ^2}}})$ as defined in Eq.(\ref{eq:2}), we propose Eq.(\ref{eq:7})

	\begin{equation}\label{eq:7}
	\mathop {\min \sum\limits_{c = 1}^C {{W_{ij}}(x_i^T{x_j})} }\limits_{s.t.{{({x_i})}_{ \in {D_S}^c}},{{({x_j})}_{ \notin {D_T}^c}}}  \Rightarrow \mathop {\max \sum\limits_{c = 1}^C { - {W_{ij}}(x_i^T{x_j})} }\limits_{s.t.{{({x_i})}_{ \in {D_S}^c}},{{({x_j})}_{ \notin {D_T}^c}}}
\end{equation}

Where  a closely aligned sample pair $({({x_i})_{ \in {D_S}^c}},{({x_j})_{ \notin {D_T}^c}})$ on the left-hand of Eq.(\ref{eq:7}) receives large weight ${{W_{ij}}}$, thereby leading to a  large constraint when reducing the value of ${x_i^T{x_j}}$, resulting in separating  the closely aligned yet differently labeled samples $({({x_i})_{ \in {D_S}^c}},{({x_j})_{ \notin {D_T}^c}})$. In the final model,  the minimization problem on the left-hand of  Eq.(\ref{eq:7}) is reformulated as the model maximization on the right-hand of Eq.(\ref{eq:7}). 

By simple mathematical operations, Eq.(\ref{eq:7}) can be reformulated as:

	\begin{equation}\label{eq:8}
	\begin{gathered}
  \max ({\mathbf{X}}({{\mathbf{G}}_{{\mathbf{SG}}}}){{\mathbf{X}}^{\mathbf{T}}}) \hfill \\
  s.t.\;\;{{\mathbf{G}}_{{\mathbf{SG}}}} =  - (1/{\mathbf{W}}).*{\mathbf{MAS}}{{\mathbf{K}}_{({n_s} + {n_t},{n_s} + {n_t})}}\; \hfill \\
  \;\;\;\;\;\;\left\{ \begin{gathered}
  \;{\mathbf{MAS}}{{\mathbf{K}}_{ij}} = 1,\;\;if\;{({{\mathbf{M}}_{S \to T}})_{ij}} = \frac{{ - 1}}{{n_s^{(c)}n_t^{(r)}}} \hfill \\
  {\mathbf{MAS}}{{\mathbf{K}}_{ij}} = 0,\;\;otherwise \hfill \\ 
\end{gathered}  \right. \hfill \\ 
\end{gathered} 
\end{equation}

where ${{\mathbf{M}}_{S \to T}}$ is defined in Eq.(\ref{eq:CDDAnew}). ${{\bf{G}}_{{\bf{SG}}}}$ defines the \textbf{Separation Graph} which formulates the different weights or attentions to drag away the closely aligned yet differently labeled cross-domain sample pairs.

\subsubsection{Final model}
\label{Final model}

Overall, the formulation of \textbf{DB-MMD} starts from \textbf{JDA}, whereof  \textbf{MMD} is leveraged to reduce the marginal/conditional distribution divergence across domains by minimizing Eq.(\ref{eq:JDA}). In making use of the specifically designed \textit{repulsive force}(\textbf{RF}) term (Eq.(\ref{eq:CDDAnew})), the \textbf{MMD} measurement is strengthened to yield the discriminative functional learning as discussed in \textbf{CDDA} \cite{luo2020discriminative}. Then, in hybridizing with Eq.(\ref{eq:5}) and Eq.(\ref{eq:8}) to enable decision boundary awareness, we formulate our final \textbf{D}ecision \textbf{B}oundary optimization-informed \textbf{MMD} (\textbf{DB-MMD}) as:

\begin{equation}\label{eq:DBMMD}
		{{\bf{M}}_{\bf{0}}} + ({{\bf{G}}_{{\bf{CG}}}}.*\sum\limits_{c = 1}^{c = C} {{{\bf{M}}_c}} ) - {{\bf{G}}_{{\bf{SG}}}}.*({{\bf{M}}_{S \to T}} + {{\bf{M}}_{T \to S}}).
\end{equation}

To highlight the benefits of the proposed \textbf{DB-MMD}, we embed it  into a baseline \textbf{DA} model, \textit{i.e.}, \textbf{CDDA} \cite{luo2020discriminative}, which only align marginal and conditional data distributions across domains as in Sect.\ref{subsection:Matching Marginal and Conditional Distributions} along with a repulsive force as defined in Sect.\ref{Repulsing interclass data for discriminative DA} for discriminative \textbf{DA}, and obtain:

\begin{equation}\label{eq:9}
		\left\{ {\begin{array}{*{20}{c}}
  {\mathop {\min }\limits_{{{\mathbf{A}}^T}{\mathbf{XH}}{{\mathbf{X}}^T}{\mathbf{A}} = {\mathbf{I}}} (tr({{\mathbf{A}}^T}{\mathbf{X}}({{\mathbf{M}}_{\mathbf{0}}} + ({{\mathbf{G}}_{{\mathbf{CG}}}}.*\sum\limits_{c = 1}^{c = C} {{{\mathbf{M}}_c}} )){{\mathbf{X}}^T}{\mathbf{A}}) + \lambda \left\| {\mathbf{A}} \right\|_F^2)} \\ 
  {\mathop {\max }\limits_{{{\mathbf{A}}^T}{\mathbf{XH}}{{\mathbf{X}}^T}{\mathbf{A}} = {\mathbf{I}}} (tr({{\mathbf{A}}^T}{\mathbf{X}}({{\mathbf{G}}_{{\mathbf{SG}}}}.*({{\mathbf{M}}_{S \to T}} + {{\mathbf{M}}_{T \to S}})){{\mathbf{X}}^T}{\mathbf{A}}) + \lambda \left\| {\mathbf{A}} \right\|_F^2)} 
\end{array}} \right.
\end{equation}

where the constraint ${{{\bf{A}}^T}{\bf{XH}}{{\bf{X}}^T}{\bf{A}} = {\bf{I}}}$ 
is derived from Principal  Component Analysis (\textbf{PCA}) to preserve the intrinsic data covariance of both domains and avoid the trivial solution for \textbf{A}. $\lambda$ is a regularization parameter which helps to well-define the optimization problem. In summary, \textbf{DB-MMD}-based \textbf{CDDA} (\textbf{CDDA+DB}) as formulated in Eq.(\ref{eq:9}) mainly focuses  on solving the following two tasks:

	\begin{itemize}
	\item \textbf{Model Minimization}: The model minimization of Eq.(\ref{eq:9}) hybridizes Eq.(\ref{eq:JDA}) and Eq.(\ref{eq:5}) by dot multiplication between ${{{\mathbf{G}}_{{\mathbf{CG}}}}}$ and $\sum {{{\mathbf{M}}_c}} $, thus giving more constraints to drag closer intra-class cross-domain samples but widely separated.

	\item \textbf{Model Maximization}: The model maximization of  Eq.(\ref{eq:9}) unifies Eq.(\ref{eq:CDDAnew}) and Eq.(\ref{eq:8}) by dot multiplication between ${({{\mathbf{M}}_{S \to T}} + {{\mathbf{M}}_{T \to S}})}$ and ${{{\mathbf{G}}_{{\mathbf{SG}}}}}$, thereby providing sufficient constraints to separate the closely aligned yet differently labeled cross-domain samples.

	\end{itemize}

    As a result, the proposed \textbf{CDDA+DB} achieves a  decision boundary aware domain adaptation while hybridizing different optimization terms in Sect.\ref{subsection:Matching Marginal and Conditional Distributions} through Sect.\ref{Alignment Graph}.

\subsection{Optimization}
\label{Optimization}

For the purpose of efficient optimization, Eq.(\ref{eq:9}) is  reformulated as Eq.(\ref{eq:10}):

\begin{equation}\label{eq:10}
		\begin{array}{c}
  \mathop {\min }\limits_{{{\mathbf{A}}^T}{\mathbf{XH}}{{\mathbf{X}}^T}{\mathbf{A}} = {\mathbf{I}}} (tr({{\mathbf{A}}^T}{\mathbf{X}}({{\mathbf{M}}_{\mathbf{0}}} + {\mathbf{DB}}){{\mathbf{X}}^T}{\mathbf{A}}) + \lambda \left\| {\mathbf{A}} \right\|_F^2) \hfill \\
  s.t.\;\;{\mathbf{DB}} = ({{\mathbf{G}}_{{\mathbf{CG}}}}.*\sum\limits_{c = 1}^{c = C} {{{\mathbf{M}}_c}} ) - ({{\mathbf{G}}_{{\mathbf{SG}}}}.*({{\mathbf{M}}_{S \to T}} + {{\mathbf{M}}_{T \to S}})) \hfill \\ 
\end{array}
\end{equation}

Optimizing Eq.(\ref{eq:10}) amounts to solving the generalized eigendecomposition problem for searching the best projection matrix ${\mathbf{A}}$. By using Augmented Lagrangian method \cite{fortin2000augmented,long2013transfer}, we obtain the best candidate matrix projection by setting its partial derivation \textit{w.r.t.} $\boldsymbol{A}$ equal to zero:

\begin{equation}\label{eq:11}
		({\bf{X}}{\bf{M_{cyd}}}{{\bf{X}}^T} + \lambda {\bf{I}}){\bf{A}} = {\bf{XH}}{{\bf{X}}^T}{\bf{A}}\Phi
\end{equation}

where $\Phi {\rm{ = diagram}}({\varphi _1},...{\varphi _k}) \in {R^{k*k}}$ is the Lagrange multiplier. Subsequently, the optimal subspace $\boldsymbol{A}$ is reduced to solving Eq.(\ref{eq:11}) for the k smallest eigenvectors. We then obtain the projection matrix $\boldsymbol{A}$ and the updated feature representation in the newly searched common feature space ${\bf{Z}} = {{\bf{A}}^T}{\bf{X}}$, thereby optimizing Eq.(\ref{eq:9}).

\subsection{Kernelization}
\label{Kernelization}

The proposed \textbf{CDDA+DB} approach is extended to nonlinear problems in a Reproducing Kernel Hilbert Space via the kernel mapping $\phi :x \to \phi (x)$, or $\phi ({\bf{X}}):[\phi ({{\bf{x}}_1}),...,\phi ({{\bf{x}}_n})]$, and the kernel matrix ${\bf{K}} = \phi {({\bf{X}})^T}\phi ({\bf{X}}) \in {R^{n*n}}$. We utilize the representer theorem to formulate the Kernel \textbf{CDDA+DB} as:

\begin{equation}\label{eq:ker}
		\begin{array}{c}
  \mathop {\min }\limits_{{{\mathbf{A}}^T}{\mathbf{XH}}{{\mathbf{X}}^T}{\mathbf{A}} = {\mathbf{I}}} (tr({{\mathbf{A}}^T}{\mathbf{K}}({{\mathbf{M}}_{\mathbf{0}}} + {\mathbf{DB}}){{\mathbf{K}}^T}{\mathbf{A}}) + \lambda \left\| {\mathbf{A}} \right\|_F^2) \hfill \\
  s.t.\;\;{\mathbf{DB}} = ({{\mathbf{G}}_{{\mathbf{CG}}}}.*\sum\limits_{c = 1}^{c = C} {{{\mathbf{M}}_c}} ) - ({{\mathbf{G}}_{{\mathbf{SG}}}}.*({{\mathbf{M}}_{S \to T}} + {{\mathbf{M}}_{T \to S}})) \hfill \\ 
	\end{array}
\end{equation}

	\section{Experiments}
    	\label{Experiments}
  Benchmarks and features are defined in Sect.\ref{subsection:Benchmarks and Features} (see Fig.\ref{fig:data}). Sect.\ref{subsection:Baseline Methods} lists the baseline methods. Sect.\ref{subsection: Experimental setup} presents the experimental setup, including  the evaluation protocol, baseline models and derived models using DB-MMD.  Sect.\ref{subsection: Experimental Results and Discussion} discusses the experimental results in comparison with the state of the art. Sect.\ref{Empirical Analysis} analyzes the convergence and parameter sensitivity of the proposed method along with a visualization of learned feature sub-spaces.    	

	\subsection{Benchmarks and Features}
    \label{subsection:Benchmarks and Features}
\begin{figure}[h!]
	\centering
	\includegraphics[width=0.7\linewidth]{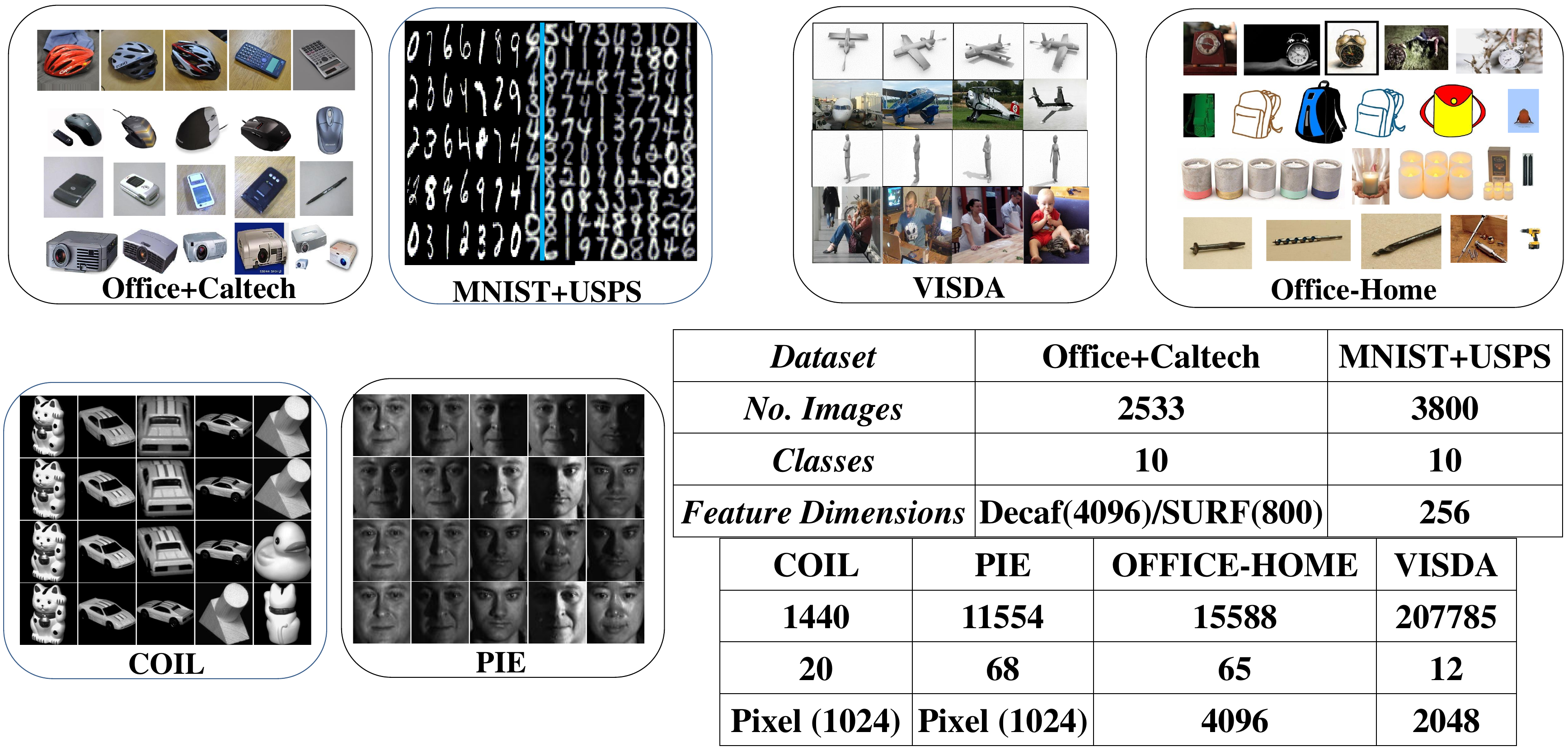}
	\caption { Sample images from 8 datasets used in our experiments. Each dataset represents a different domain. The OFFICE dataset contains three sub-datasets: DSLR, Amazon, and Webcam.} 
	\label{fig:data}	
\end{figure}

As illustrated in Fig.\ref{fig:data}, USPS\cite{DBLP:journals/pami/Hull94}+MINIST\cite{lecun1998gradient}, COIL20\cite{long2013transfer}, PIE\cite{long2013transfer}, office+Caltech\cite{long2013transfer}, Office-Home\cite{venkateswara2017deep} and VisDA\cite{peng2017visda} are standard benchmarks for  evaluation and comparison with state-of-the-art in DA. In this paper, we construct 60 datasets for different image classification tasks following the data preparation as most previous  works \cite{DBLP:journals/tcyb/UzairM17,DBLP:journals/tip/XuFWLZ16,DBLP:conf/icml/GongGS13,DBLP:journals/pami/GhifaryBKZ17,DBLP:journals/tip/DingF17,DBLP:journals/corr/LuoWHC17} do. Due
to space limitations, a detailed experimental discussion of the benchmarks and features can be found in the supplemental materials.

\subsection{Baseline Methods}
\label{subsection:Baseline Methods}

The proposed \textbf{DB-MMD} is embedded into a series of baseline approaches as defined in Sect.\ref{Baseline models}. They are  compared with  \textbf{fourty-three} methods categorized into shallow \textbf{DA} methods or deep ones:

	\begin{itemize}
		\item \textbf{Shallow methods}:
		(1) 1-Nearest Neighbor Classifier(\textbf{NN}); 
		(2) Principal Component Analysis (\textbf{PCA}); 
		(3) \textbf{GFK} \cite{gong2012geodesic}; 
		(4) \textbf{TCA} \cite{pan2011domain}; 
		(5) \textbf{TSL} \cite{4967588}; 
		(6) \textbf{JDA} \cite{long2013transfer}; 
		(7) \textbf{ELM} \cite{DBLP:journals/tcyb/UzairM17}; 
		(8) \textbf{AELM} \cite{DBLP:journals/tcyb/UzairM17}; 
		(9) \textbf{SA} \cite{DBLP:conf/iccv/FernandoHST13}; 
		(10) \textbf{mSDA} \cite{DBLP:journals/corr/abs-1206-4683}; 
		(11) \textbf{TJM} \cite{DBLP:conf/cvpr/LongWDSY14}; 
		(12) \textbf{RTML} \cite{DBLP:journals/tip/DingF17}; 
		(13) \textbf{SCA} \cite{DBLP:journals/pami/GhifaryBKZ17}; 
		(14) \textbf{CDML} \cite{DBLP:conf/aaai/WangWZX14}; 
		(15) \textbf{LTSL} \cite{DBLP:journals/ijcv/ShaoKF14}; 
		(16) \textbf{LRSR} \cite{DBLP:journals/tip/XuFWLZ16}; 
		(17) \textbf{KPCA} \cite{DBLP:journals/neco/ScholkopfSM98}; 
		(18) \textbf{JGSA}  \cite{Zhang_2017_CVPR}; 
		(19) \textbf{CORAL}  \cite{sun2016return};
		(20) \textbf{RVDLR}  \cite{jhuo2012robust};
		(21) \textbf{LPJT}  \cite{li2019locality};
		(22) \textbf{DGA-DA} \cite{luo2020discriminative};
		(23) \textbf{GEF} through its different  variants, \textbf{GEF-PCA},\textbf{GEF-LDA}, \textbf{GEF-LMNN}, and \textbf{GEF-MFA} \cite{DBLP:journals/tip/ChenS0W20};
		(24) \textbf{ARG-DA} \cite{luo2022attention};  
		(25) \textbf{DOLL-DA} \cite{luo2023discriminative}.

		\item \textbf{Deep methods}: 
		(26) \textbf{AlexNet}  \cite{krizhevsky2012imagenet};
		(27) \textbf{ResNet} \cite{he2016deep};
		(28) \textbf{ADDA} \cite{tzeng2017adversarial};
		(29) \textbf{LTRU}) \cite{sener2016learning};
		(30) \textbf{ATU} \cite{DBLP:conf/icml/SaitoUH17};
		(31) \textbf{BSWD} \cite{rozantsev2018beyond};
		(32) \textbf{DSN} \cite{bousmalis2016domain};	
		(33) \textbf{DDC} \cite{DBLP:journals/corr/TzengHZSD14};
		(34) \textbf{DAN}  \cite{long2015learning};
		(35) \textbf{TADA} \cite{wang2019transferable};
		(36) \textbf{GSDA} \cite{hu2020unsupervised};
		(37) \textbf{ETD} \cite{li2020enhanced};	
		(38) \textbf{CDAN} \cite{DBLP:conf/nips/LongC0J18};
		(39) \textbf{MCD} \cite{DBLP:conf/cvpr/SaitoWUH18};
		(40) \textbf{DAH}  \cite{venkateswara2017deep}; 
		(41) \textbf{DANN}  \cite{ganin2016domain};
  		(42) \textbf{SRDA} \cite{cai2021learning};
		(43) \textbf{MEDM} \cite{wu2021entropy}.

	\end{itemize}
	
	Direct comparison of the proposed \textbf{DB-MMD} enforced \textbf{DA} approaches using shallow features against  \textbf{DL}-based DA methods could be unfair.  Following the previous experimental settings as reported in \textbf{DGA-DA}, \textbf{JGSA}, \textbf{MCD} and \textbf{BSWD}, we made use of deep features, \textit{i.e.}, \textbf{DeCAF6} and \textbf{Resnet-50}, as the input features for fair comparison with the  \textbf{DL}-based DA methods. Whenever possible,  the reported performance scores of the \textbf{fourty-three} methods of the literature are directly  collected from their original papers or previous research  \cite{tzeng2017adversarial,DBLP:journals/tcyb/UzairM17,li2019locality,DBLP:journals/pami/GhifaryBKZ17,rozantsev2018beyond,Zhang_2017_CVPR,luo2020discriminative,DBLP:journals/tip/ChenS0W20}. They are assumed to be their \emph{best} performance.

\subsection{Experimental Setup}
\label{subsection: Experimental setup}

Evaluation protocol is first defined in (Sect.\ref{Evaluation protocol}), then the baseline models and their reinforced models using \textbf{DB-MMD} in (Sect.\ref{Baseline models}), and finally the hyper-parameter settings in (Sect.\ref{Hyper-parameter Settings}).

\subsubsection{Evaluation Protocol}
\label{Evaluation protocol}

In this research, experimental results are evaluated based on {\emph{accuracy}} of the test dataset as defined by Eq.(\ref{eq:accuracy}). It is widely used in literature, \textit{e.g.}, \cite{long2015learning,DBLP:journals/corr/LuoWHC17,long2013transfer,DBLP:journals/tip/XuFWLZ16}, \textit{etc}.
							
	\begin{equation}\label{eq:accuracy}
		\begin{array}{c}
	Accuracy = \frac{{\left| {x:x \in {D_T} \wedge \hat y(x) = y(x)} \right|}}{{\left| {x:x \in {D_T}} \right|}}
	\end{array}
	\end{equation}
where ${\cal{D_T}}$ is the target domain treated as test data, ${\hat{y}(x)}$ is the predicted label and ${y(x)}$ is the ground truth label for a test data  $x$.

\subsubsection{Baseline Models and Derived Models}
\label{Baseline models}

\begin{figure}[h!]
	\centering
	\includegraphics[width=0.7\linewidth]{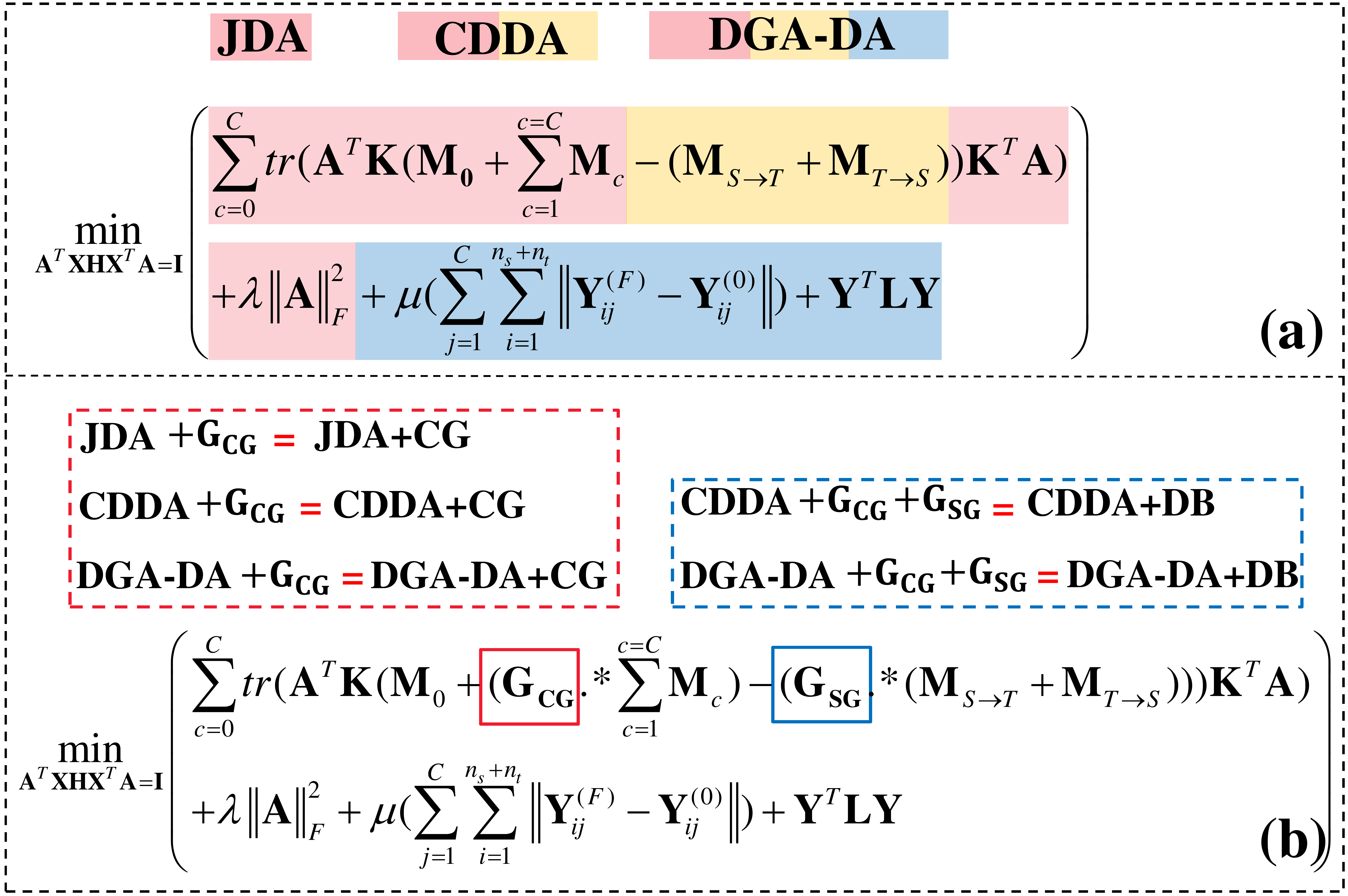}

	\caption {In Fig.\ref{fig:DB}.(a), the red portion denotes the baseline model of \textbf{JDA}, which is further improved by \textbf{CDDA} in hybridizing the \textit{repulse force} term formalized in the yellow part. Then, based on \textbf{CDDA}, a geometric regularization is also embedded to formalize the  \textbf{DGA-DA}. Fig.\ref{fig:DB}.(b) illustrates the derived \textbf{DA} models based on the three baseline models in Fig.\ref{fig:DB}.(a).}
		\label{fig:DB}
\end{figure}

We have so far proposed the \textbf{compacting} (${{\mathbf{G}}_{{\mathbf{CG}}}}$) and \textbf{separation graph} (${{\mathbf{G}}_{{\mathbf{SG}}}}$) as in Eq.(\ref{eq:5}) and Eq.(\ref{eq:8}), respectively, to make the \textbf{MMD} measurement aware of decision boundaries. In order to highlight their effectiveness and hybridization,  four baseline models were selected, namely, \textbf{JDA}, \textbf{CDDA}, \textbf{DGA-DA}, and \textbf{MEDA},  \textit{w.r.t}, their  variants , \textit{e.g.}, \textbf{JDA+CG}, \textbf{CDDA+CG}, \textbf{DGA-DA+CG}, \textbf{MEDA+CG},  \textbf{CDDA+DB} and \textbf{DGA-DA+DB}, whereof \textbf{'+CG'} represents the baseline  method  hybridized with the \textbf{compacting graph}, while \textbf{'+DB'} denotes that the baseline \textbf{DA} method reinforced with both the \textbf{compacting} and  \textbf{separation graph}.  Fig.\ref{fig:DB} details the mathematical formulations of these reinforced  \textbf{DA} methods.


\begin{itemize}
	\item \textbf{Baseline DA methods}:
	\end{itemize}

\textbf{JDA$ \Rightarrow $CDDA$ \Rightarrow $DGA-DA}: In Fig.\ref{fig:DB}.(a), \textbf{JDA} is mathematically formalized in the pink areas. It only makes use of Eq.(\ref{eq:JDA}) to shrink the marginal/conditional distribution divergence. Subsequently, \textbf{CDDA} improves \textbf{JDA} by additionally introducing the \textit{repulse force} term (Eq.(\ref{eq:CDDAnew})) to bring in data   discriminativeness  across the different domains. Based on \textbf{CDDA}, \textbf{DGA-DA} further unifies the cross-domain labeling functions through manifold learning in optimizing the Label Smoothness Consistency (\textbf{LSmC}) term \cite{lazarou2021iterative,luo2020discriminative}.

 \textbf{MEDA}: This \textbf{DA} model also improves \textbf{JDA} by exploring the hidden data geometric knowledge through minimizing $tr({{\mathbf{A}}^{\mathbf{T}}}{\mathbf{K}}(\rho {\mathbf{L}}){\mathbf{KA}})$. Moreover,  the pre-defined labels on the source domain are also used to train a regression model in generating the pseudo labels. As a result,  \textbf{MEDA} is formalized as:

\begin{equation}\label{eq:14}
		\min \left( \begin{gathered}
  tr({{\mathbf{A}}^{\mathbf{T}}}{\mathbf{K}}(\alpha {\mathbf{M}} + \rho {\mathbf{L}}){\mathbf{KA}}) + \eta tr({{\mathbf{A}}^{\mathbf{T}}}{\mathbf{KA}}) \hfill \\
   + \left\| {({\mathbf{Y}} - {{\mathbf{A}}^{\mathbf{T}}}{\mathbf{K}}){\mathbf{A}}} \right\|_F^2 \hfill \\ 
\end{gathered}  \right)
\end{equation}

\begin{itemize}
	\item \textbf{Derived data boundary aware DA models}:
	\end{itemize}

\textbf{JDA+CG, CDDA+CG, DGA-DA+CG, MEDA+CG}: In these experimental settings, the proposed \textit{Compacting Graph (CG)}  is hybridized with the baseline models, \textit{i.e.}, \textbf{JDA, CDDA, DGA-DA}, and \textbf{MEDA}, to make them compacting intra-class instances, and generate the \textbf{CG} enforced \textbf{DA} methods. The former three derived \textbf{DA} models are mathematically depicted in Fig.\ref{fig:DB}.(b), while  \textbf{MEDA+CG} is based on Eq.(\ref{eq:14}) and thus formulated as:

\begin{equation}\label{eq:15}
		\min \left( \begin{gathered}
  tr({{\mathbf{A}}^{\mathbf{T}}}{\mathbf{K}}(\alpha ({\mathbf{M}}.*{{\mathbf{G}}_{{\mathbf{CG}}}}) + \rho {\mathbf{L}}){\mathbf{KA}}) \hfill \\
   + \eta tr({{\mathbf{A}}^{\mathbf{T}}}{\mathbf{KA}}) + \left\| {({\mathbf{Y}} - {{\mathbf{A}}^{\mathbf{T}}}{\mathbf{K}}){\mathbf{A}}} \right\|_F^2 \hfill \\ 
\end{gathered}  \right)
\end{equation}
These partial \textbf{DB} models thus enable to quantify the benefits of adding the \textbf{CG} constraint into the original \textbf{DB} models.

\textbf{CDDA+DB, DGA-DA+DB}: Because  \textbf{CDDA} and \textbf{DGA-DA} integrate already in their model a \textit{repulsive force} term in contrast to \textbf{MEDA}, we can further embed the \textit{Separation Graph (SG)} into the previous derived \textbf{DA} models, \textit{i.e.}, \textbf{CDDA+CG}, and \textbf{DGA-DA+CG}, excluding \textbf{MEDA}, to make them aware of inter-class samples,  we achieve the full data boundary reinforced \textbf{DA} models and generate  the \textbf{CDDA+DB}, and \textbf{DGA-DA+DB} models. These two full data boundary aware \textbf{DA} methods thus enable to quantify the additional benefits  of the \textit{Separation Graph} by comparing it with the \textbf{CDDA+CG}, and \textbf{DGA-DA+CG} models, and highlight the effectiveness of the proposed \textbf{DB-MMD} by comparing it with their baseline models, \textit{i.e.}, \textbf{CDDA}, and \textbf{DGA-DA} methods, respectively.

\subsubsection{Hyper-parameter Settings}
\label{Hyper-parameter Settings}

Given the fact that the target domain has no labeled data under the experimental setting of \textbf{UDA}, it is therefore not possible to tune a set of optimal hyper-parameters. Following the setting of previous research \cite{luo2020discriminative,long2013transfer,DBLP:journals/tip/XuFWLZ16}, we also evaluate the proposed \textbf{DA} approaches by empirically searching in the parameter space for the \emph{optimal} settings. Specifically, the derived \textbf{JDA+CG}, \textbf{CDDA+CG} and \textbf{CDDA+DB} methods have two hyper-parameters, \textit{i.e.}, the subspace dimension $k$, and regularization parameters $\lambda $. In our experiments, we set $k = 100$ and 1) $\lambda= 0.1$ for \textbf{USPS}, \textbf{MNIST}, \textbf{COIL20}, and \textbf{PIE}, 2) $\lambda  = 1$ for \textbf{Office+Caltech}, and \textbf{Office+HOME}. The derived \textbf{DGA-DA+CG} and \textbf{DGA-DA+DB} methods have three hyper-parameters, \textit{i.e.}, the subspace dimension $k$, regularization parameters $\lambda $ and  $\mu $. In our experiments, we set $k = 100$, $\mu = 0.01$ and 1) $\lambda= 0.1$ for \textbf{USPS}, \textbf{MNIST}, \textbf{COIL20}, and \textbf{PIE}, 2) $\lambda  = 1$ for \textbf{Office+Caltech}, and \textbf{Office+HOME}. The derived \textbf{MEDA+CG} method have four hyper-parameters, \textit{i.e.}, the subspace dimension $k$, regularization parameters $\alpha$, $\rho$, and $\eta$. In our experiments, we set $\alpha = 10$, $\rho  = 0.1$, and $\eta= 1$ and 1) $k = 20$ for \textbf{Office+Caltech}, 2) $k = 100$ for \textbf{PIE}, and \textbf{Office+HOME}. Additionally, to ensure a fair comparison, we adopt the same hyper-parameter settings for kernel configurations as the baseline methods \cite{luo2020discriminative,long2013transfer,wang2018visual}.

\subsection{Experimental Results and Discussion}
\label{subsection: Experimental Results and Discussion}

\subsubsection{\textbf{Experiments on the CMU PIE Dataset}}
\label{subsubsection:Experiments on CMU PIE dataset}
\begin{figure*}[h!]
	\centering
	\includegraphics[width=1\linewidth]{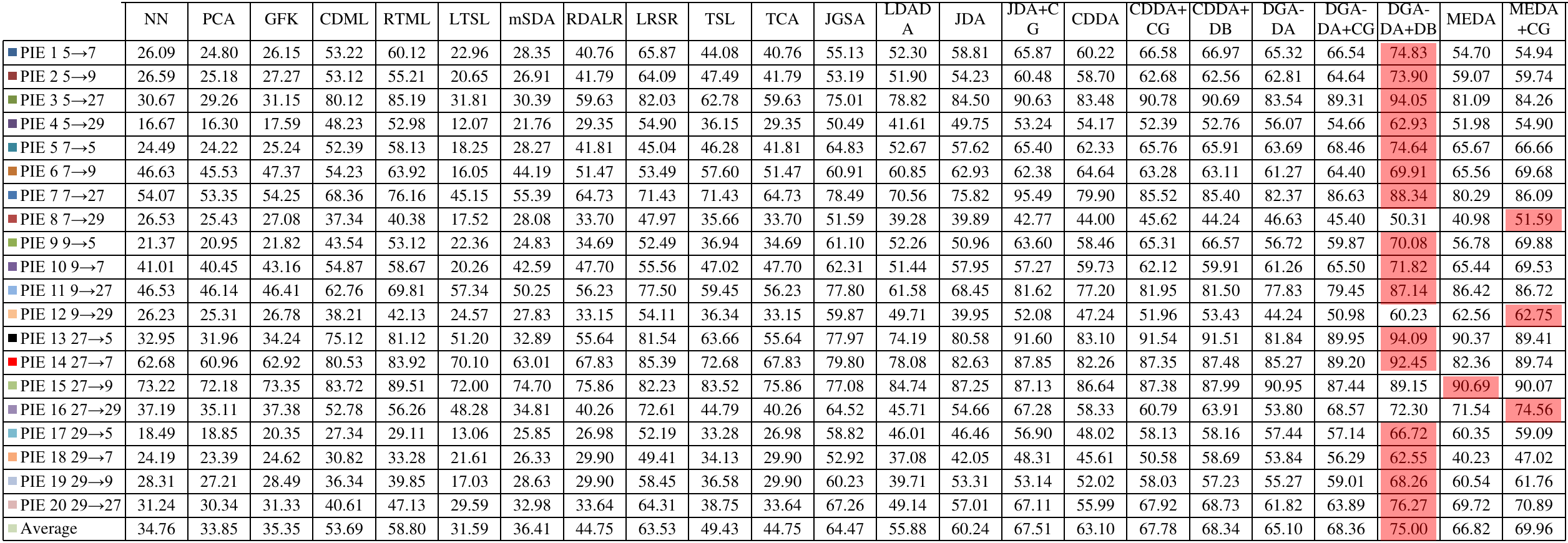}

	\caption {Accuracy${\rm{\% }}$ on the PIE Images Dataset.}
		\label{fig:PIE}
\end{figure*}

The \textbf{CMU PIE} database is a large face dataset consisting  68 people with different pose, illumination, and variations in expression. Fig.\ref{fig:PIE} synthesizes the experimental results of the various \textbf{DA} method using this dataset with the top results  highlighted in red color.

As can be seen in Fig.\ref{fig:PIE}, by integrating the decision boundary (DB) awareness, the proposed reinforced models, \textit{i.e.}, \textbf{DGA-DA+DB} and \textbf{MEDA+CG}, significantly improve over their baseline models, \textit{i.e.}, \textbf{DGA-DA} and \textbf{MEDA}, by ${\bf 9.5}\uparrow$ and ${\bf 3.1}\uparrow$ points, respectively. Specifically, in integrating the compacting graph (CG), the derived \textbf{JDA+CG}, \textbf{CDDA+CG}, \textbf{DGA-DA+CG}, and \textbf{MEDA+CG} approaches generally improve over their baseline models, \textit{i.e.}, \textbf{JDA}, \textbf{CDDA}, \textbf{DGA-DA}, and \textbf{MEDA}, by ${\bf 7.5}\uparrow$, ${\bf 4.6}\uparrow$, ${\bf 3.2}\uparrow$ and ${\bf 3.1}\uparrow$ points, respectively, thereby validating the effectiveness of the designed \textit{compacting graph (CG)} (Eq.(\ref{eq:5})) in aligning the cross-domain samples for the decision boundary optimization. In further integrating  the \textit{separation graph (SG)} (Eq.(\ref{eq:8})) for more comprehensive decision boundary optimization, the derived \textbf{CDDA+DB} and \textbf{DGA-DA+DB} models further lift the performance of \textbf{CDDA+CG} and \textbf{DGA-DA+CG} models by ${\bf 3.2}\uparrow$ and ${\bf 3.1}\uparrow$ points, respectively, thereby validating the decision boundary aware \textbf{DA}.

\subsubsection{\textbf{Experiments on the Office-Home Dataset}}
\label{subsubsection:Experiments on the Office-Home dataset}

\begin{figure}[h!]
	\centering
	\includegraphics[width=0.75\linewidth]{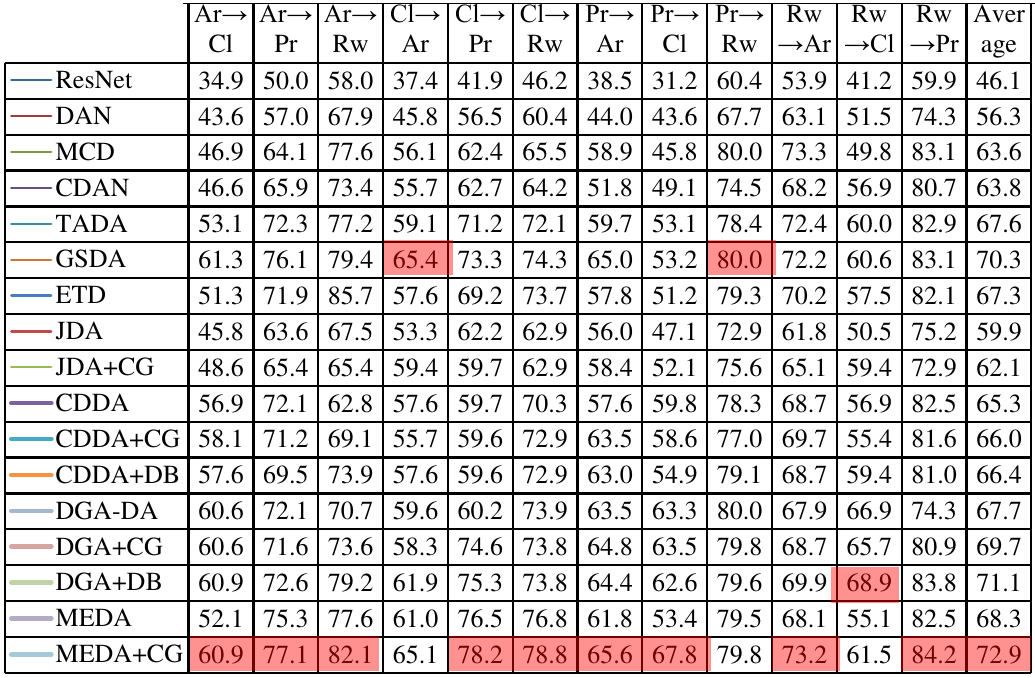}
	\caption { Accuracy${\rm{\% }}$ on the Office-Home Images Dataset.} 
	\label{fig:accHOME}
\end{figure}

As introduced in \textbf{DAH}\cite{venkateswara2017deep}, \textbf{Office-Home}  is a novel challenging benchmark for the DA task. It contains 4 different domains with 65 object categories, thereby generating 12 different DA tasks. Fig.\ref{fig:accHOME} synthesizes the performance of the proposed DA methods with  \textbf{Resnet-50} features.

As can be seen in  Fig.\ref{fig:accHOME}, without any explicit \textit{distribution divergence reduction}, the baseline method \textbf{ResNet} achieves $44.48\%$ accuracy. By reducing the domain shift through multi-kernel \textbf{MMD} measurements, \textbf{DAN} lifts the performance and achieves $56.3\%$ average accuracy. \textbf{CDAN}\cite{DBLP:conf/nips/LongC0J18}  further proposes the conditional adversarial mechanism to shrink the conditional domain divergence, and receives $7.4$ points improvement as compared with \textbf{DAN}. \textbf{MCD} \cite{DBLP:conf/cvpr/SaitoWUH18} embraces the adversarial trick using two classifiers and the feature generator for  decision boundary optimization, and achieves $63.6\%$ average accuracy. \textbf{ETD} \cite{li2020enhanced} calculates the transport distances between the cross-domain samples to further strengthen the \textbf{DA} model and achieves $67.3\%$ average accuracy. In hybridizing attention regularized functional learning and hierarchical gradient synchronization, the deep learning based \textbf{DA} methods, \textbf{TADA} \cite{wang2019transferable} and \textbf{GSDA} \cite{hu2020unsupervised}, display $67.6\%$ and $70.3\%$ accuracy, respectively. Interestingly, with $71.1\%$ and $72.9\%$ accuracy, the proposed \textbf{MEDA+CG} and \textbf{DGA-DA+DB} achieve the first and second best performance, and outperform a series of \textbf{DL} based DA methods, suggesting the competitiveness of the proposed decision boundary aware \textbf{DA} methods.

Specifically, in Fig.\ref{fig:accHOME}, the derived \textbf{JDA+CG}, \textbf{CDDA+CG}, \textbf{DGA-DA+CG}, and \textbf{MEDA+CG} models  improve once again over their baseline models, \textit{i.e.}, \textbf{JDA}, \textbf{CDDA}, \textbf{DGA-DA}, and \textbf{MEDA}, by ${\bf 2.2}\uparrow$, ${\bf 0.7}\uparrow$, ${\bf 2.0}\uparrow$, and ${\bf 4.6}\uparrow$ points, respectively, thereby suggesting the contribution of the proposed \textit{Compacting graph} (Eq.(\ref{eq:5})) in optimizing the decision boundary for an  effective \textbf{DA}. Now, by further leveraging the \textit{separation graph} (Eq.(\ref{eq:8})) for more comprehensive decision boundary optimization, the derived \textbf{CDDA+DB} and \textbf{DGA-DA+DB} models further lift the \textbf{DA} performance over their only \textbf{CG}-based counterparts, \textit{i.e.}, \textbf{CDDA+CG}, \textbf{DGA-DA+CG}, by ${\bf 0.4}\uparrow$, ${\bf 1.4}\uparrow$,  and achieve with $66.4\%$ and $77.1\%$ accuracy , thereby suggesting the usefulness of  the \textbf{DB-MMD} enforced decision boundary aware \textbf{DA}.

\subsubsection{\textbf{Experiments on the Office+Caltech-256 Data Sets}}
\label{subsubsection:Experiments on the Office+Caltech-256 Data Sets}

Due to space limitations, the experimental results can be found in the supplemental materials.

\subsubsection{\textbf{Experiments on the COIL 20 Dataset}} 
\label{subsubsection: results on the COIL dataset}
Due to space limitations, the experimental results can be found in the supplemental materials.

\subsubsection{\textbf{Experiments on the USPS+MNIST Data Set}}
\label{subsubsection: experiments on the UPS+MNIST Datasets}
Due to space limitations, the experimental results can be found in the supplemental materials.

\subsubsection{\textbf{Experiments on the VISDA Data Set}}
\label{subsubsection: experiments on the VISDA Datasets}
Due to space limitations, the experimental results can be found in the supplemental materials.

\subsubsection{\textbf{Discussion}}
\label{subsubsection: Discussion}
As can be seen in the previous subsections, the derived \textbf{DB-MMD} enhanced \textbf{DA} models enable to improve the performance of their baseline models, \textit{i.e.}, \textbf{JDA}, \textbf{CDDA}, \textbf{DGA-DA} and \textbf{MEDA}, and even display state of the art performance over 60 \textbf{DA} tasks through 8 datasets. While the proposed reinforced models, \textit{e.g.}, \textbf{DGA-DA+DB} and \textbf{MEDA+CG}, significantly improve over their baseline models on \textbf{CMU PIE}, \textit{i.e.}, \textbf{DGA-DA} and \textbf{MEDA}, by ${\bf 9.5}\uparrow$ and ${\bf 3.1}\uparrow$ points, respectively, they only improve slightly or are in par with their baseline models on other datasets, including the \textbf{Office+Caltech-256, COIL 20 and USPS+MNIST}, but however never harm the performance of the baseline models. The reason of such a behavior will be subject of our future investigation.

\subsection{Empirical Analysis}
\label{Empirical Analysis}

An important question of the proposed \textbf{DB-MMD} enforced \textbf{DA} models   is its sensitivity \textit{w.r.t.} the different hyper-parameter settings  (Sect.\ref{feature dimension}, Sect.\ref{over-fitting regularization}, Sect.\ref{manifold learning} ) as well as how fast the derived models converge (sect.\ref{Convergence analysis}). In sect.\ref{graph construction}, t-SNE visualization experiments were proposed to vividly quantify the effectiveness the designed \textbf{DB-MMD} in optimizing the decision boundary for more discriminative functional learning.

\subsubsection{\textbf{Sensitivity of the proposed \textbf{DB-MMD} \textit{w.r.t.} to feature dimension}}
\label{feature dimension} 

The feature dimension $k$ denotes the dimension of the searched shared latent feature subspace, which determines the structure of low-dimension embedding. Obviously, the larger $k$ is  the better the shared subspace can afford complex data distributions, but at the cost of increased computation complexity. In this section, our research objective is to observe the sensitivity of the designed \textit{compacting graph} and \textit{separation graph} \textit{w.r.t.} the feature dimension by comparing the baseline models and their variants.

\begin{figure}[h!]
	\centering
	\includegraphics[width=1\linewidth]{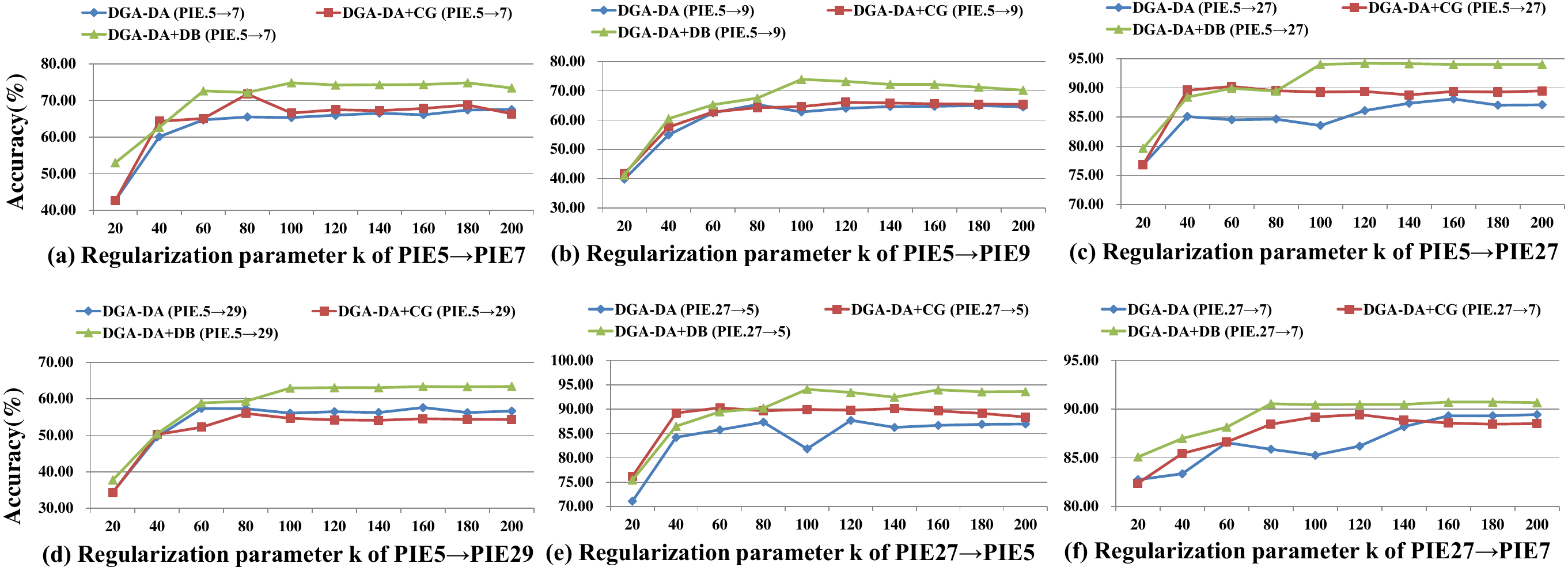}

	\caption {Sensitivity analysis of the proposed methods, \textit{i.e.}, \textbf{DGA-DA}, \textbf{DGA-DA+CG} and \textbf{DGA-DA+DB},  using PIE dataset \textit{w.r.t.} subspace dimension  $k$.}
		\label{fig:k}
\end{figure}

In Fig.\ref{fig:k}, using \textbf{PIE} dataset, 6 cross-domain tasks were proposed ( \emph{PIE.5} $\rightarrow$ \emph{PIE.7} ...  \emph{PIE.27} $\rightarrow$ \emph{PIE.7}  ) to observe the performance of the  baseline model, \textit{i.e.}, \textbf{DGA-DA}, and its derived models (\textbf{DGA-DA+CG} and \textbf{DGA-DA+DB}) by changing feature dimension $k$. As shown in Fig.\ref{fig:k}, the subspace dimensionality $k$ varies with $k \in \{20,40,60,80,100,150,200\}$. Both derived \textbf{DA} methods, \textit{i.e.}, \textbf{DGA-DA+CG} and \textbf{DGA-DA+DB}, remain stable \textit{w.r.t.}  a wide range of  $k \in \{ 100 \le k  \le 200\} $.

Interestingly, in Fig.\ref{fig:k}.(c) and Fig.\ref{fig:k}.(f), the performance of \textbf{DGA-DA} doesn't  achieve stability until $k$ reaches 160, while the derived models, \textit{i.e.}, \textbf{DGA-DA+CG} and \textbf{DGA-DA+DB}, stopped improvement once $k$ reaches $100$, suggesting that the designed \textit{compacting graph} and \textit{separation graph} enforced \textbf{DA} methods enjoy prompt optimization in the low dimensional feature space,  result in  efficient \textbf{DA} methods.

\subsubsection{\textbf{Sensitivity of the proposed \textbf{DB-MMD} \textit{w.r.t.} to over-fitting regularization}}
\label{over-fitting regularization} 

Over-fitting regularization is important for model optimization, which reduces the risk of the model's over-fitting, but shrinks the diverse representation of the functional learning. In \textbf{DGA-DA} and the derived \textbf{DA} models, \textit{i.e.}, \textbf{DGA-DA+CG} and \textbf{DGA-DA+DB}, $\lambda$ is the designed hyper-parameter which regularizes the projection matrix ${\bf{A}}$, thereby avoiding over-fitting the chosen shared feature subspace \textit{w.r.t.} both source and target domain.

\begin{figure}[h!]
	\centering
	\includegraphics[width=1\linewidth]{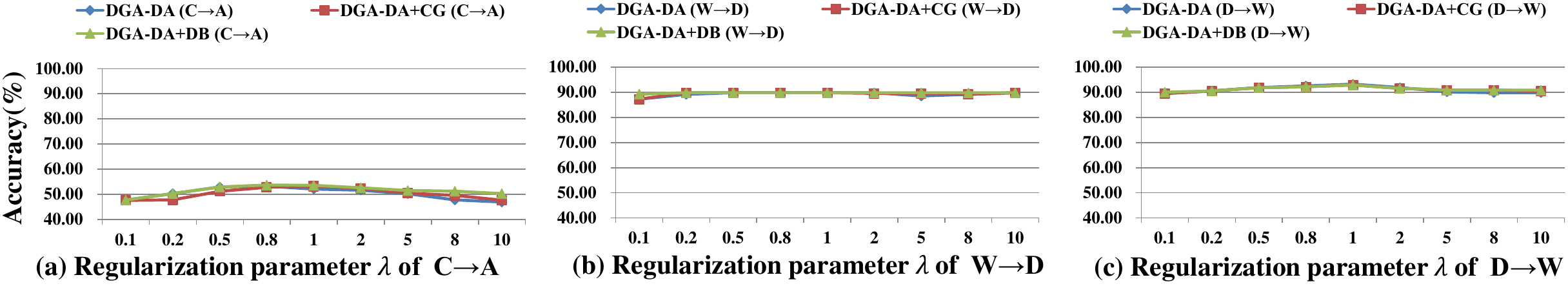}

	\caption {The classification accuracies of the proposed \textbf{DGA-DA}, \textbf{DGA-DA+CG} and \textbf{DGA-DA+DB} methods vs. the parameter $\lambda $ on the selected three cross domains data sets.}
		\label{fig:lambda}
\end{figure}

In Fig.\ref{fig:lambda}, we plot the classification accuracy of the proposed \textbf{DA} methods \textit{w.r.t} different values of $\lambda$ on the \textbf{Office+Caltech} datasets using the \textbf{DeCAF6} features. As shown in Fig.\ref{fig:lambda}, the hyper-parameter $\lambda$ varies with $\lambda  \in \{0.1,0.2,0.5,0.8,1,2,5,8,10\}$, yet the baseline model \textbf{DGA-DA} and its variants, \textit{i.e.}, \textbf{DGA-DA+CG} and \textbf{DGA-DA+DB}, remain stable \textit{w.r.t.}  a wide range of with $\lambda \in \{ 0.1 \le k  \le 10\} $,  suggesting that the proposed methods can easily search the proper candidate for the over-fitting hyper-parameter.

\subsubsection{\textbf{Sensitivity of the proposed \textbf{DB-MMD} \textit{w.r.t.} to manifold learning}}
\label{manifold learning}

\begin{figure}[h!]
	\centering
	\includegraphics[width=0.7\linewidth]{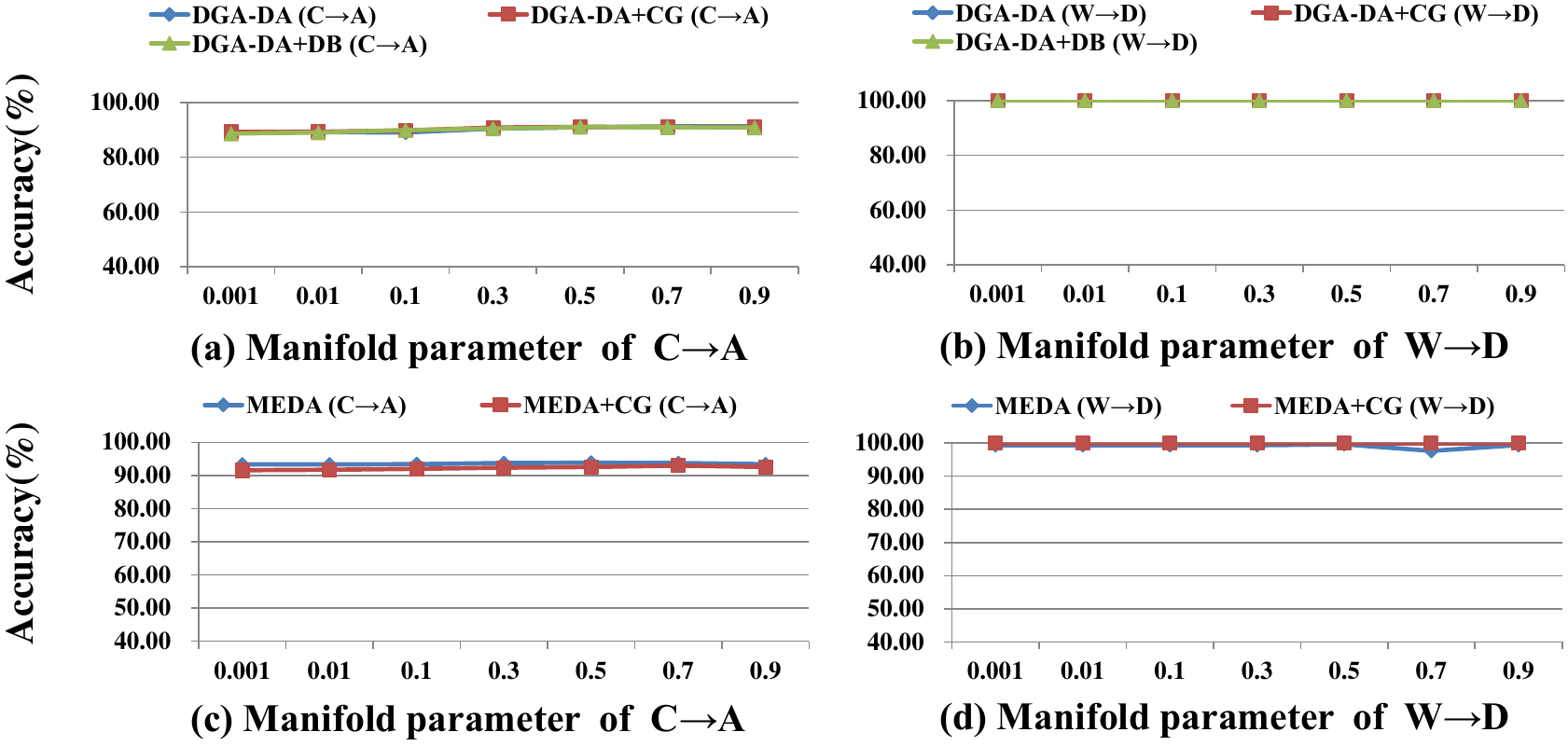}

	\caption {Detailed discussion of the proposed decision boundary aware mechanism \textit{w.r.t} different manifold learning strategies.}
		\label{fig:manifold}
\end{figure}

Manifold learning techniques have been widely applied in \textbf{DA} algorithms for lifting the performance in solving cross-domain tasks, while preserving the geometric structure across different domains and as a result  unifying the cross-domain classifiers. On the other side, the proposed decision boundary aware mechanism intends to update the geometric structure for more discriminative functional learning.  It would therefore be interesting to discuss the robustness of different experimental settings of the manifold learning in hybridizing the proposed decision boundary aware optimization strategies. For this purpose, we propose Fig.\ref{fig:manifold} to further explore the sensitivity of the proposed \textbf{DB-MMD} \textit{w.r.t.} different manifold learning settings. In Fig.\ref{fig:manifold}, using SURF features of the Office+Caltech-256 dataset, we plot the results on cross-domain tasks (\emph{C} $\rightarrow$ \emph{A}, \emph{W} $\rightarrow$ \emph{D}) based on the baseline manifold learning enforced \textbf{DA} methods and their variant models. For a comprehensive discussion, we introduce two typical yet different manifold learning strategies:

\begin{itemize}

	\item \textbf{Manifold learning across feature space and label space}:  \textbf{DGA-DA} and its variants, \textit{i.e.}, \textbf{DGA-DA+CG} and \textbf{DGA-DA+DB}, align the manifold structure across the optimized common feature space and the label space by minimizing Eq.(\ref{eq:16}):
	
	\begin{equation}\label{eq:16}
	\min (\mu (\sum\limits_{j = 1}^C {\sum\limits_{i = 1}^{{n_s} + {n_t}} {\left\| {{\mathbf{Y}}_{ij}^{(F)} - {\mathbf{Y}}_{ij}^{(0)}} \right\|} } ) + {{\mathbf{Y}}^T}{\mathbf{LY}}).
\end{equation}
	
	In the latter optimization \cite{luo2020discriminative}, $\alpha {\rm{ = }}\frac{1}{{1 + \mu }}$ is the trade-off parameter to balance the effectiveness of the smooth label propagation (	${\left\| {{\mathbf{Y}}_{ij}^{(F)} - {\mathbf{Y}}_{ij}^{(0)}} \right\|}$) and the manifold learning (${{\mathbf{Y}}^T}{\mathbf{LY}}$). Increasing $\alpha$ therefore improves the effectiveness of manifold regularization.
	
	\item \textbf{Manifold learning across different label spaces}: As depicted in  Eq.(\ref{eq:14}) and Eq.(\ref{eq:15}), both \textbf{MEDA} and  \textbf{MEDA-AG} align the manifold structure of the original feature space and the optimized common feature space by minimizing $tr({{\mathbf{A}}^{\mathbf{T}}}{\mathbf{K}}(\rho {\mathbf{L}}){\mathbf{KA}})$. Thus, increasing $\rho $ improves the effectiveness of manifold regularization.

\end{itemize}

As shown in Fig.\ref{fig:manifold},  the manifold learning parameters $\alpha$ and $\rho$ vary with $\{ \alpha ,\rho \}  \in \{ 0.001,\\ 0.01,0.1,0.3,0.5,0.7,0.9\}$, yet the baseline models and their variants remain stable \textit{w.r.t.}  a wide range of $\{ \alpha ,\rho \}  \in \{ 0.0001 \leqslant \{ \alpha ,\rho \}  \prec 1\} $. The results therefore suggest that the designed \textit{compacting graph} and \textit{separation graph} enforced \textbf{DA} models are robust \textit{w.r.t} different manifold alignment strategies and can easily search the best candidate hyper-parameters,

\subsubsection{\textbf{Convergence analysis}}
\label{Convergence analysis}

\begin{figure}[h!]
	\centering
	\includegraphics[width=1\linewidth]{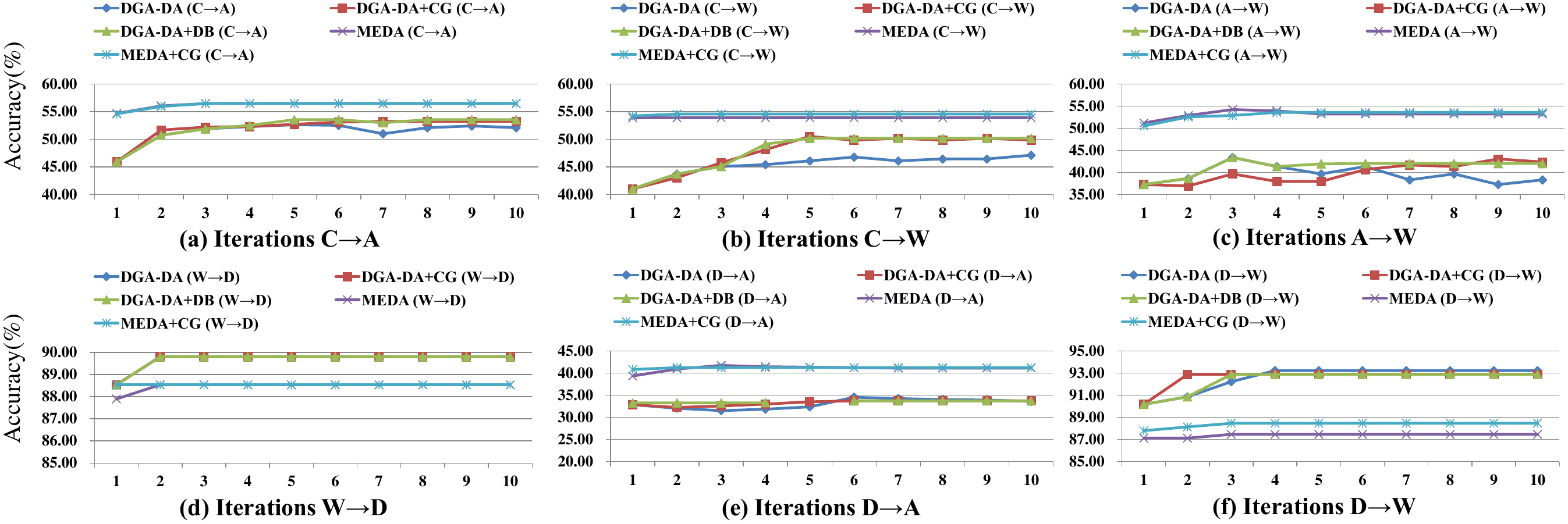}

	\caption {Convergence analysis using 6 cross-domain image classification tasks on the \textbf{Office+Caltech} dataset. (accuracy w.r.t $\#$iterations)}
		\label{fig:iteration}
\end{figure}

Another interesting question is whether the proposed decision boundary aware mechanism enforced \textbf{DA} methods enjoy the efficient model convergence. For this propose, in Fig.\ref{fig:iteration}, we perform  convergence analysis of the baseline models \textbf{DGA-DA} and \textbf{MEDA}, and their derived models, \textit{i.e.}, \textbf{DGA-DA+CG}, \textbf{DGA-DA+DB}, and \textbf{MEDA+CG}, using the \textbf{SURF} features on the \textbf{Office+Caltech} datasets. Subsequently, Fig.\ref{fig:iteration} reports 6 cross domain adaptation experiments ( \emph{C} $\rightarrow$ \emph{A}, \emph{C} $\rightarrow$ \emph{W} ... \emph{D} $\rightarrow$ \emph{A} , \emph{D} $\rightarrow$ \emph{W}  ) with the number of iterations $T = (1,2,3,4,5,6,7,8,9,10)$.  As observed in Fig.\ref{fig:iteration}, all the baseline and  derived models converge within 5$ \sim $6 iterations when performing model optimization over different datasets. Thanks to the decision boundary aware mechanism, the convergence curves displayed by \textbf{DGA-DA+CG}, \textbf{DGA-DA+DB} are more flat  in  comparison with \textbf{DGA-DA}.

\subsubsection{\textbf{t-SNE Visualization}}
\label{graph construction}

\begin{figure}[h!]
	\centering
	\includegraphics[width=0.8\linewidth]{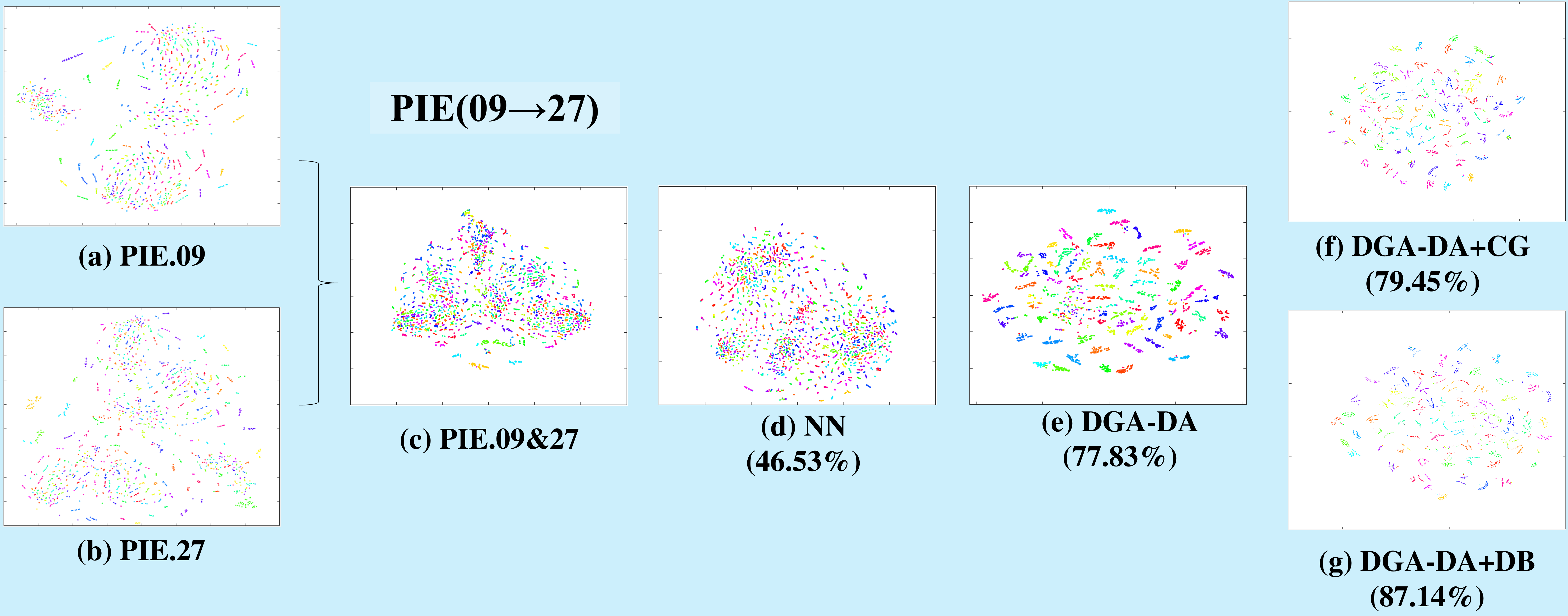}

	\caption {Accuracy(${\rm{\% }}$) and Visualization results of the \emph{PIE-9} $\rightarrow$ \emph{PIE-27} \textbf{DA} task. Fig.\ref{fig:r34}(a), Fig.\ref{fig:r34}(b), and Fig.\ref{fig:r34}(c) are visualization results of \emph{PIE-9}, \emph{PIE-27}, and \emph{PIE-27\&9} datasets in their \textbf{Original} data space, respectively. Fig.\ref{fig:r34}(d) visualizes both the source and target datasets after \textbf{NN} classification without distribution divergence reduction. Subsequently, after domain adaptation, Fig.\ref{fig:r34}(e), Fig.\ref{fig:r34}(f), and Fig.\ref{fig:r34}(g) visualize both the source and target datasets in \textbf{DGA-DA}, \textbf{DGA-DA+CG}, and \textbf{DGA-DA+DB} subspaces, respectively. The 68 facial classes are represented in different colors.}
		\label{fig:r34}
\end{figure}

Using the PIE dataset and t-SNE visualization \cite{van2008visualizing} method, Fig.\ref{fig:r34} visualizes the class explicit data distributions in their original subspace and the resultant shared feature subspace, respectively, using the proposed \textbf{DGA+CG}, \textbf{DGA+DB} and their baseline model, \textit{i.e.}, \textbf{DGA-DA}, under similar experimental setting. Additionally, the visualization results were reported using the basic classifier without distribution divergence reduction, \textit{i.e.}, Nearest Neighbor (\textbf{NN}), to highlight the effectiveness of the proposed methods in reducing domain divergence.

\begin{itemize}
	
	\item \textbf{Data distributions and geometric structures}: Fig.\ref{fig:r34}(a,b,c) visualizes the \emph{PIE-9}, \emph{PIE-27},  and \emph{PIE-9}$\& $\emph{PIE-27} datasets in their \textbf{Original} data space, respectively. As can be observed  from these figures, the samples from each of the 68  sub-domains or classes, colored differently according to each sub-domain, are randomly and disorderly displayed in their low dimensional embedded feature spaces.

	\item \textbf{Baseline model without \textbf{DA}}: Fig.\ref{fig:r34}(d) makes use of the base classifier, \textit{i.e.} \textbf{NN}, to classify the target domain samples, which are very confused between classes across the domains due to the significant cross-domain divergence, leading to very poor performance.

	\item \textbf{\textbf{DA} enforced cross-domain classification}:  \textbf{DGA-DA} explicitly explores the discriminative statistic distribution alignment and the manifold structure regularization, thereby significantly improving \textbf{NN}'s classification accuracy by ${\bf 31.30}\uparrow$  points. Based on \textbf{DGA-DA}, adding the	designed \textit{compacting graph} (Eq.(\ref{eq:5})) to further optimize the decision boundary for more effective \textbf{DA}, \textbf{DGA-DA+CG}, as illustrated in Fig.\ref{fig:r34}(f), shows a better separation of the different sub-domains in the \textbf{DGA-DA+CG}'s embedding feature space. Lastly, as shown in Fig.\ref{fig:r34}(g), our final model \textbf{DGA-DA+DB} further improves  \textbf{DGA-DA+CG} by ${\bf 7.69}\uparrow$ points, thereby highlighting the importance of the  designed \textit{separation graph} (Eq.(\ref{eq:8})) and demonstrating the effectiveness of the designed \textbf{DB-MMD} in searching the discriminative functional learning for \textbf{DA}.

\end{itemize}

\section{Conclusion}

In this paper, a novel distribution measurement, Decision Boundary optimization-informed Maximum Mean Discrepancy (\textbf{DB-MMD}), has been proposed. It enables the cross-domain distribution measurement with the ability to specifically detect the  samples at decision boundaries, thus generating the decision boundary optimized functional learning. We have specifically designed \textbf{compacting graph} to shrink the divergence among the cross-domain samples with same labels. Meanwhile, we have also explored the discriminative effectiveness within  the different labeled cross sub-domains by using the designed \textbf{separation graph}. For a comprehensive discussion,  four popular \textbf{DA} methods were selected as the baseline models to hybridize the designed \textbf{compacting} and \textbf{separation graph} terms, respectively. Different \textbf{DA} models were then derived to quantify the contribution of the proposed decision boundary aware mechanism.   Using  real data, alongside six variants of the proposed \textbf{DA} methods,  we have further provided in-depth analysis and insight into the proposed \textbf{DB-MMD}, in quantifying and visualizing the contribution of the decision boundary optimization and data discriminativeness.    Future works include a better understanding of the behavior differences of the proposed \textbf{DB-MMD} via  various datasets, and   embedding of the proposed \textbf{DB-MMD}   into the paradigm of deep learning in order to improve various computer vision applications, \textit{e.g.}, detection, segmentation, and tracking, \textit{etc}.

\vskip 0.2in
\bibliography{sample}

\begin{thebibliography}{85}
\providecommand{\natexlab}[1]{#1}
\providecommand{\url}[1]{\texttt{#1}}
\expandafter\ifx\csname urlstyle\endcsname\relax
  \providecommand{\doi}[1]{doi: #1}\else
  \providecommand{\doi}{doi: \begingroup \urlstyle{rm}\Url}\fi

\bibitem[Belkin and Niyogi(2003)]{belkin2003laplacian}
Mikhail Belkin and Partha Niyogi.
\newblock Laplacian eigenmaps for dimensionality reduction and data representation.
\newblock \emph{Neural computation}, 15\penalty0 (6):\penalty0 1373--1396, 2003.

\bibitem[Ben-David et~al.(2010)Ben-David, Blitzer, Crammer, Kulesza, Pereira, and Vaughan]{ben2010theory}
Shai Ben-David, John Blitzer, Koby Crammer, Alex Kulesza, Fernando Pereira, and Jennifer~Wortman Vaughan.
\newblock A theory of learning from different domains.
\newblock \emph{Machine learning}, 79\penalty0 (1):\penalty0 151--175, 2010.

\bibitem[Borgwardt et~al.(2006)Borgwardt, Gretton, Rasch, Kriegel, Sch{\"o}lkopf, and Smola]{borgwardt2006integrating}
Karsten~M Borgwardt, Arthur Gretton, Malte~J Rasch, Hans-Peter Kriegel, Bernhard Sch{\"o}lkopf, and Alex~J Smola.
\newblock Integrating structured biological data by kernel maximum mean discrepancy.
\newblock \emph{Bioinformatics}, 22\penalty0 (14):\penalty0 e49--e57, 2006.

\bibitem[Bousmalis et~al.(2016)Bousmalis, Trigeorgis, Silberman, Krishnan, and Erhan]{bousmalis2016domain}
Konstantinos Bousmalis, George Trigeorgis, Nathan Silberman, Dilip Krishnan, and Dumitru Erhan.
\newblock Domain separation networks.
\newblock In \emph{Advances in Neural Information Processing Systems}, pages 343--351, 2016.

\bibitem[Cai et~al.(2021)Cai, He, Zhou, Alhumade, and Hu]{cai2021learning}
Guanyu Cai, Lianghua He, MengChu Zhou, Hesham Alhumade, and Die Hu.
\newblock Learning smooth representation for unsupervised domain adaptation.
\newblock \emph{IEEE Transactions on Neural Networks and Learning Systems}, 2021.

\bibitem[Chen et~al.(2012)Chen, Xu, Weinberger, and Sha]{DBLP:journals/corr/abs-1206-4683}
Minmin Chen, Zhixiang~Eddie Xu, Kilian~Q. Weinberger, and Fei Sha.
\newblock Marginalized denoising autoencoders for domain adaptation.
\newblock \emph{CoRR}, abs/1206.4683, 2012.
\newblock URL \url{http://arxiv.org/abs/1206.4683}.

\bibitem[Chen et~al.(2020)Chen, Song, Li, and Wu]{DBLP:journals/tip/ChenS0W20}
Yiming Chen, Shiji Song, Shuang Li, and Cheng Wu.
\newblock A graph embedding framework for maximum mean discrepancy-based domain adaptation algorithms.
\newblock \emph{{IEEE} Trans. Image Process.}, 29:\penalty0 199--213, 2020.
\newblock \doi{10.1109/TIP.2019.2928630}.
\newblock URL \url{https://doi.org/10.1109/TIP.2019.2928630}.

\bibitem[Cho et~al.(2022)Cho, Kim, Jung, and Kweon]{cho2022mcdal}
Jae~Won Cho, Dong-Jin Kim, Yunjae Jung, and In~So Kweon.
\newblock Mcdal: Maximum classifier discrepancy for active learning.
\newblock \emph{IEEE Transactions on Neural Networks and Learning Systems}, 2022.

\bibitem[Courty et~al.(2017{\natexlab{a}})Courty, Flamary, Habrard, and Rakotomamonjy]{courty2017joint}
Nicolas Courty, R{\'e}mi Flamary, Amaury Habrard, and Alain Rakotomamonjy.
\newblock Joint distribution optimal transportation for domain adaptation.
\newblock In \emph{Advances in Neural Information Processing Systems}, pages 3733--3742, 2017{\natexlab{a}}.

\bibitem[Courty et~al.(2017{\natexlab{b}})Courty, Flamary, Tuia, and Rakotomamonjy]{courty2017optimal}
Nicolas Courty, R{\'e}mi Flamary, Devis Tuia, and Alain Rakotomamonjy.
\newblock Optimal transport for domain adaptation.
\newblock \emph{IEEE transactions on pattern analysis and machine intelligence}, 39\penalty0 (9):\penalty0 1853--1865, 2017{\natexlab{b}}.

\bibitem[Ding and Fu(2017)]{DBLP:journals/tip/DingF17}
Zhengming Ding and Yun Fu.
\newblock Robust transfer metric learning for image classification.
\newblock \emph{{IEEE} Trans. Image Processing}, 26\penalty0 (2):\penalty0 660--670, 2017.
\newblock \doi{10.1109/TIP.2016.2631887}.
\newblock URL \url{https://doi.org/10.1109/TIP.2016.2631887}.

\bibitem[Fernando et~al.(2013)Fernando, Habrard, Sebban, and Tuytelaars]{DBLP:conf/iccv/FernandoHST13}
Basura Fernando, Amaury Habrard, Marc Sebban, and Tinne Tuytelaars.
\newblock Unsupervised visual domain adaptation using subspace alignment.
\newblock In \emph{Proceedings of the IEEE international conference on computer vision}, pages 2960--2967, 2013.

\bibitem[Fortin and Glowinski(2000)]{fortin2000augmented}
Michel Fortin and Roland Glowinski.
\newblock \emph{Augmented Lagrangian methods: applications to the numerical solution of boundary-value problems}, volume~15.
\newblock Elsevier, 2000.

\bibitem[Ganin et~al.(2016)Ganin, Ustinova, Ajakan, Germain, Larochelle, Laviolette, Marchand, and Lempitsky]{ganin2016domain}
Yaroslav Ganin, Evgeniya Ustinova, Hana Ajakan, Pascal Germain, Hugo Larochelle, Fran{\c{c}}ois Laviolette, Mario Marchand, and Victor Lempitsky.
\newblock Domain-adversarial training of neural networks.
\newblock \emph{The Journal of Machine Learning Research}, 17\penalty0 (1):\penalty0 2096--2030, 2016.

\bibitem[Ghifary et~al.(2017)Ghifary, Balduzzi, Kleijn, and Zhang]{DBLP:journals/pami/GhifaryBKZ17}
Muhammad Ghifary, David Balduzzi, W.~Bastiaan Kleijn, and Mengjie Zhang.
\newblock Scatter component analysis: {A} unified framework for domain adaptation and domain generalization.
\newblock \emph{{IEEE} Trans. Pattern Anal. Mach. Intell.}, 39\penalty0 (7):\penalty0 1414--1430, 2017.
\newblock \doi{10.1109/TPAMI.2016.2599532}.
\newblock URL \url{https://doi.org/10.1109/TPAMI.2016.2599532}.

\bibitem[Gong et~al.(2012)Gong, Shi, Sha, and Grauman]{gong2012geodesic}
Boqing Gong, Yuan Shi, Fei Sha, and Kristen Grauman.
\newblock Geodesic flow kernel for unsupervised domain adaptation.
\newblock In \emph{Computer Vision and Pattern Recognition (CVPR), 2012 IEEE Conference on}, pages 2066--2073. IEEE, 2012.

\bibitem[Gong et~al.(2013)Gong, Grauman, and Sha]{DBLP:conf/icml/GongGS13}
Boqing Gong, Kristen Grauman, and Fei Sha.
\newblock Connecting the dots with landmarks: Discriminatively learning domain-invariant features for unsupervised domain adaptation.
\newblock In \emph{International conference on machine learning}, pages 222--230. PMLR, 2013.

\bibitem[Goodfellow et~al.(2014)Goodfellow, Pouget-Abadie, Mirza, Xu, Warde-Farley, Ozair, Courville, and Bengio]{goodfellow2014generative}
Ian Goodfellow, Jean Pouget-Abadie, Mehdi Mirza, Bing Xu, David Warde-Farley, Sherjil Ozair, Aaron Courville, and Yoshua Bengio.
\newblock Generative adversarial nets.
\newblock In \emph{Advances in neural information processing systems}, pages 2672--2680, 2014.

\bibitem[Goodfellow et~al.(2016)Goodfellow, Bengio, and Courville]{goodfellow2016deep}
Ian Goodfellow, Yoshua Bengio, and Aaron Courville.
\newblock \emph{Deep learning}.
\newblock MIT press, 2016.

\bibitem[Gretton et~al.(2006)Gretton, Borgwardt, Rasch, Sch{\"o}lkopf, and Smola]{DBLP:conf/nips/GrettonBRSS06}
Arthur Gretton, Karsten Borgwardt, Malte Rasch, Bernhard Sch{\"o}lkopf, and Alex Smola.
\newblock A kernel method for the two-sample-problem.
\newblock \emph{Advances in neural information processing systems}, 19, 2006.

\bibitem[He et~al.(2016)He, Zhang, Ren, and Sun]{he2016deep}
Kaiming He, Xiangyu Zhang, Shaoqing Ren, and Jian Sun.
\newblock Deep residual learning for image recognition.
\newblock In \emph{Proceedings of the IEEE conference on computer vision and pattern recognition}, pages 770--778, 2016.

\bibitem[Heo et~al.(2019)Heo, Lee, Yun, and Choi]{heo2019knowledge}
Byeongho Heo, Minsik Lee, Sangdoo Yun, and Jin~Young Choi.
\newblock Knowledge distillation with adversarial samples supporting decision boundary.
\newblock In \emph{Proceedings of the AAAI conference on artificial intelligence}, volume~33, pages 3771--3778, 2019.

\bibitem[Herath et~al.(2017)Herath, Harandi, and Porikli]{herath2017learning}
Samitha Herath, Mehrtash Harandi, and Fatih Porikli.
\newblock Learning an invariant hilbert space for domain adaptation.
\newblock In \emph{Proceedings of the IEEE Conference on Computer Vision and Pattern Recognition}, pages 3845--3854, 2017.

\bibitem[Hoffman et~al.(2018)Hoffman, Tzeng, Park, Zhu, Isola, Saenko, Efros, and Darrell]{pmlr-v80-hoffman18a}
Judy Hoffman, Eric Tzeng, Taesung Park, Jun-Yan Zhu, Phillip Isola, Kate Saenko, Alexei Efros, and Trevor Darrell.
\newblock {C}y{CADA}: Cycle-consistent adversarial domain adaptation.
\newblock In Jennifer Dy and Andreas Krause, editors, \emph{Proceedings of the 35th International Conference on Machine Learning}, volume~80 of \emph{Proceedings of Machine Learning Research}, pages 1989--1998, Stockholmsmässan, Stockholm Sweden, 10--15 Jul 2018. PMLR.
\newblock URL \url{http://proceedings.mlr.press/v80/hoffman18a.html}.

\bibitem[Hu et~al.(2020)Hu, Kan, Shan, and Chen]{hu2020unsupervised}
Lanqing Hu, Meina Kan, Shiguang Shan, and Xilin Chen.
\newblock Unsupervised domain adaptation with hierarchical gradient synchronization.
\newblock In \emph{Proceedings of the IEEE/CVF Conference on Computer Vision and Pattern Recognition}, pages 4043--4052, 2020.

\bibitem[Hull(1994)]{DBLP:journals/pami/Hull94}
Jonathan~J. Hull.
\newblock A database for handwritten text recognition research.
\newblock \emph{{IEEE} Trans. Pattern Anal. Mach. Intell.}, 16\penalty0 (5):\penalty0 550--554, 1994.
\newblock \doi{10.1109/34.291440}.
\newblock URL \url{https://doi.org/10.1109/34.291440}.

\bibitem[Jhuo et~al.(2012)Jhuo, Liu, Lee, and Chang]{jhuo2012robust}
I-Hong Jhuo, Dong Liu, DT~Lee, and Shih-Fu Chang.
\newblock Robust visual domain adaptation with low-rank reconstruction.
\newblock In \emph{Computer Vision and Pattern Recognition (CVPR), 2012 IEEE Conference on}, pages 2168--2175. IEEE, 2012.

\bibitem[Karbalayghareh et~al.(2018)Karbalayghareh, Qian, and Dougherty]{karbalayghareh2018optimal}
Alireza Karbalayghareh, Xiaoning Qian, and Edward~R Dougherty.
\newblock Optimal bayesian transfer learning.
\newblock \emph{IEEE Transactions on Signal Processing}, 66\penalty0 (14):\penalty0 3724--3739, 2018.

\bibitem[Kifer et~al.(2004)Kifer, Ben-David, and Gehrke]{kifer2004detecting}
Daniel Kifer, Shai Ben-David, and Johannes Gehrke.
\newblock Detecting change in data streams.
\newblock In \emph{Proceedings of the Thirtieth international conference on Very large data bases-Volume 30}, pages 180--191. VLDB Endowment, 2004.

\bibitem[Kim et~al.(2019)Kim, Sahu, Gholami, and Pavlovic]{kim2019unsupervised}
Minyoung Kim, Pritish Sahu, Behnam Gholami, and Vladimir Pavlovic.
\newblock Unsupervised visual domain adaptation: A deep max-margin gaussian process approach.
\newblock In \emph{Proceedings of the IEEE/CVF Conference on Computer Vision and Pattern Recognition}, pages 4380--4390, 2019.

\bibitem[Krizhevsky et~al.(2012)Krizhevsky, Sutskever, and Hinton]{krizhevsky2012imagenet}
Alex Krizhevsky, Ilya Sutskever, and Geoffrey~E Hinton.
\newblock Imagenet classification with deep convolutional neural networks.
\newblock In \emph{Advances in neural information processing systems}, pages 1097--1105, 2012.

\bibitem[Lazarou et~al.(2021)Lazarou, Stathaki, and Avrithis]{lazarou2021iterative}
Michalis Lazarou, Tania Stathaki, and Yannis Avrithis.
\newblock Iterative label cleaning for transductive and semi-supervised few-shot learning.
\newblock In \emph{Proceedings of the IEEE/CVF International Conference on Computer Vision}, pages 8751--8760, 2021.

\bibitem[LeCun et~al.(1998)LeCun, Bottou, Bengio, and Haffner]{lecun1998gradient}
Yann LeCun, L{\'e}on Bottou, Yoshua Bengio, and Patrick Haffner.
\newblock Gradient-based learning applied to document recognition.
\newblock \emph{Proceedings of the IEEE}, 86\penalty0 (11):\penalty0 2278--2324, 1998.

\bibitem[Lee et~al.(2021)Lee, Cho, and Im]{lee2021dranet}
Seunghun Lee, Sunghyun Cho, and Sunghoon Im.
\newblock Dranet: Disentangling representation and adaptation networks for unsupervised cross-domain adaptation.
\newblock In \emph{Proceedings of the IEEE/CVF Conference on Computer Vision and Pattern Recognition}, pages 15252--15261, 2021.

\bibitem[Li et~al.(2021{\natexlab{a}})Li, Li, Shi, and Yu]{li2021cross}
Jichang Li, Guanbin Li, Yemin Shi, and Yizhou Yu.
\newblock Cross-domain adaptive clustering for semi-supervised domain adaptation.
\newblock In \emph{Proceedings of the IEEE/CVF Conference on Computer Vision and Pattern Recognition}, pages 2505--2514, 2021{\natexlab{a}}.

\bibitem[Li et~al.(2019)Li, Jing, Lu, Zhu, and Shen]{li2019locality}
Jingjing Li, Mengmeng Jing, Ke~Lu, Lei Zhu, and Heng~Tao Shen.
\newblock Locality preserving joint transfer for domain adaptation.
\newblock \emph{IEEE Transactions on Image Processing}, 28\penalty0 (12):\penalty0 6103--6115, 2019.

\bibitem[Li et~al.(2020{\natexlab{a}})Li, Chen, Ding, Zhu, Lu, and Shen]{li2020maximum}
Jingjing Li, Erpeng Chen, Zhengming Ding, Lei Zhu, Ke~Lu, and Heng~Tao Shen.
\newblock Maximum density divergence for domain adaptation.
\newblock \emph{IEEE transactions on pattern analysis and machine intelligence}, 43\penalty0 (11):\penalty0 3918--3930, 2020{\natexlab{a}}.

\bibitem[Li et~al.(2020{\natexlab{b}})Li, Zhai, Luo, Ge, and Ren]{li2020enhanced}
Mengxue Li, Yi-Ming Zhai, You-Wei Luo, Peng-Fei Ge, and Chuan-Xian Ren.
\newblock Enhanced transport distance for unsupervised domain adaptation.
\newblock In \emph{Proceedings of the IEEE/CVF Conference on Computer Vision and Pattern Recognition}, pages 13936--13944, 2020{\natexlab{b}}.

\bibitem[Li et~al.(2021{\natexlab{b}})Li, Lv, Xie, Liu, Liang, and Qin]{li2021bi}
Shuang Li, Fangrui Lv, Binhui Xie, Chi~Harold Liu, Jian Liang, and Chen Qin.
\newblock Bi-classifier determinacy maximization for unsupervised domain adaptation.
\newblock In \emph{AAAI}, 2021{\natexlab{b}}.

\bibitem[Liang et~al.(2021)Liang, Hu, and Feng]{liang2021domain}
Jian Liang, Dapeng Hu, and Jiashi Feng.
\newblock Domain adaptation with auxiliary target domain-oriented classifier.
\newblock In \emph{Proceedings of the IEEE/CVF Conference on Computer Vision and Pattern Recognition}, pages 16632--16642, 2021.

\bibitem[Long et~al.(2013)Long, Wang, Ding, Sun, and Yu]{long2013transfer}
Mingsheng Long, Jianmin Wang, Guiguang Ding, Jiaguang Sun, and Philip~S Yu.
\newblock Transfer feature learning with joint distribution adaptation.
\newblock In \emph{Proceedings of the IEEE International Conference on Computer Vision}, pages 2200--2207, 2013.

\bibitem[Long et~al.(2014{\natexlab{a}})Long, Wang, Ding, Pan, and Yu]{DBLP:journals/tkde/LongWDPY14}
Mingsheng Long, Jianmin Wang, Guiguang Ding, Sinno~Jialin Pan, and Philip~S. Yu.
\newblock Adaptation regularization: {A} general framework for transfer learning.
\newblock \emph{{IEEE} Trans. Knowl. Data Eng.}, 26\penalty0 (5):\penalty0 1076--1089, 2014{\natexlab{a}}.
\newblock \doi{10.1109/TKDE.2013.111}.
\newblock URL \url{https://doi.org/10.1109/TKDE.2013.111}.

\bibitem[Long et~al.(2014{\natexlab{b}})Long, Wang, Ding, Sun, and Yu]{DBLP:conf/cvpr/LongWDSY14}
Mingsheng Long, Jianmin Wang, Guiguang Ding, Jiaguang Sun, and Philip~S Yu.
\newblock Transfer joint matching for unsupervised domain adaptation.
\newblock In \emph{Proceedings of the IEEE conference on computer vision and pattern recognition}, pages 1410--1417, 2014{\natexlab{b}}.

\bibitem[Long et~al.(2015)Long, Cao, Wang, and Jordan]{long2015learning}
Mingsheng Long, Yue Cao, Jianmin Wang, and Michael~I Jordan.
\newblock Learning transferable features with deep adaptation networks.
\newblock In \emph{ICML}, pages 97--105, 2015.

\bibitem[Long et~al.(2017)Long, Zhu, Wang, and Jordan]{DBLP:conf/icml/LongZ0J17}
Mingsheng Long, Han Zhu, Jianmin Wang, and Michael~I Jordan.
\newblock Deep transfer learning with joint adaptation networks.
\newblock In \emph{International conference on machine learning}, pages 2208--2217. PMLR, 2017.

\bibitem[Long et~al.(2018)Long, Cao, Wang, and Jordan]{DBLP:conf/nips/LongC0J18}
Mingsheng Long, Zhangjie Cao, Jianmin Wang, and Michael~I. Jordan.
\newblock Conditional adversarial domain adaptation.
\newblock In \emph{Neural Information Processing Systems 2018, NeurIPS 2018,}, pages 1647--1657, 2018.
\newblock URL \url{https://proceedings.neurips.cc/paper/2018/hash/ab88b15733f543179858600245108dd8-Abstract.html}.

\bibitem[Lu et~al.(2018)Lu, Shen, Cao, Xiao, and van~den Hengel]{lu2018embarrassingly}
Hao Lu, Chunhua Shen, Zhiguo Cao, Yang Xiao, and Anton van~den Hengel.
\newblock An embarrassingly simple approach to visual domain adaptation.
\newblock \emph{IEEE Transactions on Image Processing}, 2018.

\bibitem[Lu et~al.(2020)Lu, Luo, Huang, Wang, and Chen]{DBLP:journals/csur/LuLHWC20}
Ying Lu, Lingkun Luo, Di~Huang, Yunhong Wang, and Liming Chen.
\newblock Knowledge transfer in vision recognition: {A} survey.
\newblock \emph{{ACM} Comput. Surv.}, 53\penalty0 (2):\penalty0 37:1--37:35, 2020.
\newblock \doi{10.1145/3379344}.
\newblock URL \url{https://doi.org/10.1145/3379344}.

\bibitem[Luo et~al.(2017)Luo, Wang, Hu, and Chen]{DBLP:journals/corr/LuoWHC17}
Lingkun Luo, Xiaofang Wang, Shiqiang Hu, and Liming Chen.
\newblock Robust data geometric structure aligned close yet discriminative domain adaptation.
\newblock \emph{CoRR}, abs/1705.08620, 2017.
\newblock URL \url{http://arxiv.org/abs/1705.08620}.

\bibitem[Luo et~al.(2020)Luo, Chen, Hu, Lu, and Wang]{luo2020discriminative}
Lingkun Luo, Liming Chen, Shiqiang Hu, Ying Lu, and Xiaofang Wang.
\newblock Discriminative and geometry-aware unsupervised domain adaptation.
\newblock \emph{IEEE Transactions on Cybernetics}, 2020.

\bibitem[Luo et~al.(2022)Luo, Chen, and Hu]{luo2022attention}
Lingkun Luo, Liming Chen, and Shiqiang Hu.
\newblock Attention regularized laplace graph for domain adaptation.
\newblock \emph{IEEE Transactions on Image Processing}, 31:\penalty0 7322--7337, 2022.

\bibitem[Luo et~al.(2023)Luo, Hu, and Chen]{luo2023discriminative}
Lingkun Luo, Shiqiang Hu, and Liming Chen.
\newblock Discriminative noise robust sparse orthogonal label regression-based domain adaptation.
\newblock \emph{International Journal of Computer Vision}, pages 1--24, 2023.

\bibitem[Ng et~al.(2002)Ng, Jordan, and Weiss]{NIPS2001_2092}
Andrew~Y. Ng, Michael~I. Jordan, and Yair Weiss.
\newblock On spectral clustering: Analysis and an algorithm.
\newblock In T.~G. Dietterich, S.~Becker, and Z.~Ghahramani, editors, \emph{Advances in Neural Information Processing Systems 14}, pages 849--856. MIT Press, 2002.
\newblock URL \url{http://papers.nips.cc/paper/2092-on-spectral-clustering-analysis-and-an-algorithm.pdf}.

\bibitem[Pan and Yang(2010)]{pan2010survey}
Sinno~Jialin Pan and Qiang Yang.
\newblock A survey on transfer learning.
\newblock \emph{IEEE Transactions on knowledge and data engineering}, 22\penalty0 (10):\penalty0 1345--1359, 2010.

\bibitem[Pan et~al.(2011)Pan, Tsang, Kwok, and Yang]{pan2011domain}
Sinno~Jialin Pan, Ivor~W Tsang, James~T Kwok, and Qiang Yang.
\newblock Domain adaptation via transfer component analysis.
\newblock \emph{IEEE Transactions on Neural Networks}, 22\penalty0 (2):\penalty0 199--210, 2011.

\bibitem[Patel et~al.(2015)Patel, Gopalan, Li, and Chellappa]{7078994}
V.~M. Patel, R.~Gopalan, R.~Li, and R.~Chellappa.
\newblock Visual domain adaptation: A survey of recent advances.
\newblock \emph{IEEE Signal Processing Magazine}, 32\penalty0 (3):\penalty0 53--69, May 2015.
\newblock ISSN 1053-5888.
\newblock \doi{10.1109/MSP.2014.2347059}.

\bibitem[Pei et~al.(2018)Pei, Cao, Long, and Wang]{pei2018multi}
Zhongyi Pei, Zhangjie Cao, Mingsheng Long, and Jianmin Wang.
\newblock Multi-adversarial domain adaptation.
\newblock In \emph{Thirty-Second AAAI Conference on Artificial Intelligence}, 2018.

\bibitem[Peng et~al.(2017)Peng, Usman, Kaushik, Hoffman, Wang, and Saenko]{peng2017visda}
Xingchao Peng, Ben Usman, Neela Kaushik, Judy Hoffman, Dequan Wang, and Kate Saenko.
\newblock Visda: The visual domain adaptation challenge.
\newblock \emph{arXiv preprint arXiv:1710.06924}, 2017.

\bibitem[Roller et~al.(2004)Roller, Taskar, and Guestrin]{roller2004max}
BTCGD Roller, C~Taskar, and D~Guestrin.
\newblock Max-margin markov networks.
\newblock \emph{Advances in neural information processing systems}, 16:\penalty0 25, 2004.

\bibitem[Rozantsev et~al.(2018)Rozantsev, Salzmann, and Fua]{rozantsev2018beyond}
Artem Rozantsev, Mathieu Salzmann, and Pascal Fua.
\newblock Beyond sharing weights for deep domain adaptation.
\newblock \emph{IEEE Transactions on Pattern Analysis and Machine Intelligence}, 2018.

\bibitem[Saito et~al.(2017)Saito, Ushiku, and Harada]{DBLP:conf/icml/SaitoUH17}
Kuniaki Saito, Yoshitaka Ushiku, and Tatsuya Harada.
\newblock Asymmetric tri-training for unsupervised domain adaptation.
\newblock In \emph{International conference on machine learning}, pages 2988--2997. PMLR, 2017.

\bibitem[Saito et~al.(2018)Saito, Watanabe, Ushiku, and Harada]{DBLP:conf/cvpr/SaitoWUH18}
Kuniaki Saito, Kohei Watanabe, Yoshitaka Ushiku, and Tatsuya Harada.
\newblock Maximum classifier discrepancy for unsupervised domain adaptation.
\newblock In \emph{2018 {IEEE} Conference on Computer Vision and Pattern Recognition, {CVPR} 2018, Salt Lake City, UT, USA, June 18-22, 2018}, pages 3723--3732. Computer Vision Foundation / {IEEE} Computer Society, 2018.
\newblock \doi{10.1109/CVPR.2018.00392}.
\newblock URL \url{http://openaccess.thecvf.com/content\_cvpr\_2018/html/Saito\_Maximum\_Classifier\_Discrepancy\_CVPR\_2018\_paper.html}.

\bibitem[Sch{\"{o}}lkopf et~al.(1998)Sch{\"{o}}lkopf, Smola, and M{\"{u}}ller]{DBLP:journals/neco/ScholkopfSM98}
Bernhard Sch{\"{o}}lkopf, Alexander~J. Smola, and Klaus{-}Robert M{\"{u}}ller.
\newblock Nonlinear component analysis as a kernel eigenvalue problem.
\newblock \emph{Neural Computation}, 10\penalty0 (5):\penalty0 1299--1319, 1998.
\newblock \doi{10.1162/089976698300017467}.
\newblock URL \url{https://doi.org/10.1162/089976698300017467}.

\bibitem[Sener et~al.(2016)Sener, Song, Saxena, and Savarese]{sener2016learning}
Ozan Sener, Hyun~Oh Song, Ashutosh Saxena, and Silvio Savarese.
\newblock Learning transferrable representations for unsupervised domain adaptation.
\newblock In \emph{Advances in Neural Information Processing Systems}, pages 2110--2118, 2016.

\bibitem[Shao et~al.(2015)Shao, Zhu, and Li]{DBLP:journals/tnn/ShaoZL15}
Ling Shao, Fan Zhu, and Xuelong Li.
\newblock Transfer learning for visual categorization: {A} survey.
\newblock \emph{{IEEE} Trans. Neural Netw. Learning Syst.}, 26\penalty0 (5):\penalty0 1019--1034, 2015.
\newblock \doi{10.1109/TNNLS.2014.2330900}.
\newblock URL \url{https://doi.org/10.1109/TNNLS.2014.2330900}.

\bibitem[Shao et~al.(2014)Shao, Kit, and Fu]{DBLP:journals/ijcv/ShaoKF14}
Ming Shao, Dmitry Kit, and Yun Fu.
\newblock Generalized transfer subspace learning through low-rank constraint.
\newblock \emph{International Journal of Computer Vision}, 109\penalty0 (1-2):\penalty0 74--93, 2014.
\newblock \doi{10.1007/s11263-014-0696-6}.
\newblock URL \url{http://dx.doi.org/10.1007/s11263-014-0696-6}.

\bibitem[Si et~al.(2010)Si, Tao, and Geng]{4967588}
S.~Si, D.~Tao, and B.~Geng.
\newblock Bregman divergence-based regularization for transfer subspace learning.
\newblock \emph{IEEE Transactions on Knowledge and Data Engineering}, 22\penalty0 (7):\penalty0 929--942, July 2010.
\newblock ISSN 1041-4347.
\newblock \doi{10.1109/TKDE.2009.126}.

\bibitem[Sun and Saenko(2016)]{sun2016deep}
Baochen Sun and Kate Saenko.
\newblock Deep coral: Correlation alignment for deep domain adaptation.
\newblock In \emph{European Conference on Computer Vision}, pages 443--450. Springer, 2016.

\bibitem[Sun et~al.(2016)Sun, Feng, and Saenko]{sun2016return}
Baochen Sun, Jiashi Feng, and Kate Saenko.
\newblock Return of frustratingly easy domain adaptation.
\newblock In \emph{AAAI}, volume~6, page~8, 2016.

\bibitem[Tzeng et~al.(2014)Tzeng, Hoffman, Zhang, Saenko, and Darrell]{DBLP:journals/corr/TzengHZSD14}
Eric Tzeng, Judy Hoffman, Ning Zhang, Kate Saenko, and Trevor Darrell.
\newblock Deep domain confusion: Maximizing for domain invariance.
\newblock \emph{CoRR}, abs/1412.3474, 2014.
\newblock URL \url{http://arxiv.org/abs/1412.3474}.

\bibitem[Tzeng et~al.(2017)Tzeng, Hoffman, Saenko, and Darrell]{tzeng2017adversarial}
Eric Tzeng, Judy Hoffman, Kate Saenko, and Trevor Darrell.
\newblock Adversarial discriminative domain adaptation.
\newblock In \emph{Computer Vision and Pattern Recognition (CVPR)}, volume~1, page~4, 2017.

\bibitem[Uzair and Mian(2017)]{DBLP:journals/tcyb/UzairM17}
Muhammad Uzair and Ajmal~S. Mian.
\newblock Blind domain adaptation with augmented extreme learning machine features.
\newblock \emph{{IEEE} Trans. Cybernetics}, 47\penalty0 (3):\penalty0 651--660, 2017.
\newblock \doi{10.1109/TCYB.2016.2523538}.
\newblock URL \url{https://doi.org/10.1109/TCYB.2016.2523538}.

\bibitem[Van~der Maaten and Hinton(2008)]{van2008visualizing}
Laurens Van~der Maaten and Geoffrey Hinton.
\newblock Visualizing data using t-sne.
\newblock \emph{Journal of machine learning research}, 9\penalty0 (11), 2008.

\bibitem[Venkateswara et~al.(2017)Venkateswara, Eusebio, Chakraborty, and Panchanathan]{venkateswara2017deep}
Hemanth Venkateswara, Jose Eusebio, Shayok Chakraborty, and Sethuraman Panchanathan.
\newblock Deep hashing network for unsupervised domain adaptation.
\newblock \emph{arXiv preprint arXiv:1706.07522}, 2017.

\bibitem[Wang et~al.(2014)Wang, Wang, Zhang, and Xu]{DBLP:conf/aaai/WangWZX14}
Hao Wang, Wei Wang, Chen Zhang, and Fanjiang Xu.
\newblock Cross-domain metric learning based on information theory.
\newblock In \emph{Proceedings of the AAAI Conference on Artificial Intelligence}, volume~28, 2014.

\bibitem[Wang et~al.(2018)Wang, Feng, Chen, Yu, Huang, and Yu]{wang2018visual}
Jindong Wang, Wenjie Feng, Yiqiang Chen, Han Yu, Meiyu Huang, and Philip~S Yu.
\newblock Visual domain adaptation with manifold embedded distribution alignment.
\newblock In \emph{2018 ACM Multimedia Conference on Multimedia Conference}, pages 402--410. ACM, 2018.

\bibitem[Wang et~al.(2020)Wang, Chen, Feng, Yu, Huang, and Yang]{wang2020transfer}
Jindong Wang, Yiqiang Chen, Wenjie Feng, Han Yu, Meiyu Huang, and Qiang Yang.
\newblock Transfer learning with dynamic distribution adaptation.
\newblock \emph{ACM Transactions on Intelligent Systems and Technology (TIST)}, 11\penalty0 (1):\penalty0 1--25, 2020.

\bibitem[Wang et~al.(2019)Wang, Li, Ye, Long, and Wang]{wang2019transferable}
Ximei Wang, Liang Li, Weirui Ye, Mingsheng Long, and Jianmin Wang.
\newblock Transferable attention for domain adaptation.
\newblock In \emph{Proceedings of the AAAI Conference on Artificial Intelligence}, volume~33, pages 5345--5352, 2019.

\bibitem[Wei et~al.(2021)Wei, Lan, Zeng, and Chen]{wei2021toalign}
Guoqiang Wei, Cuiling Lan, Wenjun Zeng, and Zhibo Chen.
\newblock Toalign: Task-oriented alignment for unsupervised domain adaptation.
\newblock \emph{arXiv preprint arXiv:2106.10812}, 2021.

\bibitem[Wu et~al.(2021)Wu, Zhang, Zhou, Yang, Zhao, and Latecki]{wu2021entropy}
Xiaofu Wu, Suofei Zhang, Quan Zhou, Zhen Yang, Chunming Zhao, and Longin~Jan Latecki.
\newblock Entropy minimization versus diversity maximization for domain adaptation.
\newblock \emph{IEEE Transactions on Neural Networks and Learning Systems}, 2021.

\bibitem[Xu et~al.(2016)Xu, Fang, Wu, Li, and Zhang]{DBLP:journals/tip/XuFWLZ16}
Yong Xu, Xiaozhao Fang, Jian Wu, Xuelong Li, and David Zhang.
\newblock Discriminative transfer subspace learning via low-rank and sparse representation.
\newblock \emph{{IEEE} Trans. Image Processing}, 25\penalty0 (2):\penalty0 850--863, 2016.
\newblock \doi{10.1109/TIP.2015.2510498}.
\newblock URL \url{https://doi.org/10.1109/TIP.2015.2510498}.

\bibitem[Yang and Soatto(2020)]{yang2020fda}
Yanchao Yang and Stefano Soatto.
\newblock Fda: Fourier domain adaptation for semantic segmentation.
\newblock In \emph{Proceedings of the IEEE/CVF Conference on Computer Vision and Pattern Recognition}, pages 4085--4095, 2020.

\bibitem[Zhang et~al.(2017)Zhang, Li, and Ogunbona]{Zhang_2017_CVPR}
Jing Zhang, Wanqing Li, and Philip Ogunbona.
\newblock Joint geometrical and statistical alignment for visual domain adaptation.
\newblock In \emph{The IEEE Conference on Computer Vision and Pattern Recognition (CVPR)}, July 2017.

\bibitem[Zhao et~al.(2019)Zhao, Li, Yue, Gu, Xu, Hu, Chai, and Keutzer]{zhao2019multi}
Sicheng Zhao, Bo~Li, Xiangyu Yue, Yang Gu, Pengfei Xu, Runbo Hu, Hua Chai, and Kurt Keutzer.
\newblock Multi-source domain adaptation for semantic segmentation.
\newblock In \emph{Advances in Neural Information Processing Systems}, pages 7285--7298, 2019.

\bibitem[Zhu et~al.(2017)Zhu, Park, Isola, and Efros]{zhu2017unpaired}
Jun-Yan Zhu, Taesung Park, Phillip Isola, and Alexei~A Efros.
\newblock Unpaired image-to-image translation using cycle-consistent adversarial networks.
\newblock In \emph{Proceedings of the IEEE international conference on computer vision}, pages 2223--2232, 2017.

\end{thebibliography}

\end{document}